\documentclass[10pt,english]{article}
\usepackage[LGR,T1]{fontenc}
\usepackage{babel}
\usepackage{float}
\usepackage{pdfpages}
\usepackage{amstext}
\usepackage{amssymb}
\usepackage{graphicx}
\usepackage[unicode=true,pdfusetitle,
 bookmarks=true,bookmarksnumbered=false,bookmarksopen=false,
 breaklinks=false,pdfborder={0 0 1},backref=false,colorlinks=false]
 {hyperref}

\makeatletter

\DeclareRobustCommand{\greektext}{%
  \fontencoding{LGR}\selectfont\def\encodingdefault{LGR}}
\DeclareRobustCommand{\textgreek}[1]{\leavevmode{\greektext #1}}
\DeclareFontEncoding{LGR}{}{}
\DeclareTextSymbol{\~}{LGR}{126}
\usepackage{amsmath}

\let\@tmp\@xfloat     
\usepackage{fixltx2e}
\let\@xfloat\@tmp    

\usepackage{graphicx}

\usepackage{cite}

\usepackage{color} 

\usepackage[labelfont=bf,labelsep=period,justification=raggedright]{caption}

\makeatletter
\renewcommand{\@biblabel}[1]{\quad#1.}
\makeatother

\date{}

\makeatother

\begin{document}
% Title must be 150 characters or less 
\begin{flushleft} 
{\Large 
\textbf{Extraction of Pharmacokinetic Evidence of Drug-drug Interactions from the Literature} 
} 
% Insert Author names, affiliations and corresponding author email. 
\\ 
Artemy Kolchinsky$^{1,2}$,  
Anália Lourenço$^{3,4}$,  
Heng-Yi Wu,$^{5}$,
Lang Li,$^{5}$,
Luis M. Rocha,$^{1,2,\ast}$ 
\\
\bf{1} School of Informatics and Computing, Indiana University, Bloomington, IN, USA
\\
\bf{2} Instituto Gulbenkian de Ciência, Oeiras, Portugal
\\
\bf{3} ESEI: Escuela Superior de Ingeniería Informática, University of Vigo, Edificio Politécnico, Campus Universitario As Lagoas s/n 32004, Ourense, Spain
\\
\bf{4} CEB - Centre of Biological Engineering, University of Minho, Campus de Gualtar, 4710-057 Braga, Portugal
\\
\bf{5} Center for Computational Biology and Bioinformatics and Department of Medical \& Molecular Genetics, Indiana University School of Medicine, Indianapolis, IN, USA
\\
$\ast$ E-mail: rocha@indiana.edu
\end{flushleft}

\sloppy

\section*{Abstract}

Drug-drug interaction (DDI) is a major cause of morbidity and mortality
and a subject of intense scientific interest. Biomedical literature
mining can aid DDI research by extracting evidence for large numbers
of potential interactions from published literature and clinical databases.
Though DDI is investigated in domains ranging in scale from intracellular
biochemistry to human populations, literature mining has not been
used to extract specific \emph{types of experimental evidence}, which
are reported differently for distinct experimental goals. We focus
on \emph{pharmacokinetic evidence} for DDI, essential for identifying
causal mechanisms of putative interactions and as input for further
pharmacological and pharmaco-epidemiology investigations. We used
manually curated corpora of PubMed abstracts and annotated sentences
to evaluate the efficacy of literature mining on two tasks: first,
identifying PubMed abstracts containing pharmacokinetic evidence of
DDIs; second, extracting sentences containing such evidence from abstracts.
We implemented a text mining pipeline and evaluated it using several
linear classifiers and a variety of feature transforms. The most important
textual features in the abstract and sentence classification tasks
were analyzed. We also investigated the performance benefits of using
features derived from PubMed metadata fields, various publicly available
named entity recognizers, and pharmacokinetic dictionaries. Several
classifiers performed very well in distinguishing relevant and irrelevant
abstracts (reaching F1\ensuremath{\approx}0.93, MCC\ensuremath{\approx}0.74,
iAUC\ensuremath{\approx}0.99) and sentences (F1\ensuremath{\approx}0.76,
MCC\ensuremath{\approx}0.65, iAUC\ensuremath{\approx}0.83). We found
that word bigram features were important for achieving optimal classifier
performance and that features derived from Medical Subject Headings
(MeSH) terms significantly improved abstract classification. We also
found that some drug-related named entity recognition tools and dictionaries
led to slight but significant improvements, especially in classification
of evidence sentences. Based on our thorough analysis of classifiers
and feature transforms and the high classification performance achieved,
we demonstrate that literature mining can aid DDI discovery by supporting
automatic extraction of specific types of experimental evidence.

\section*{Introduction}

Drug-drug interaction (DDI) is one of the major causes of adverse
drug reaction (ADR) and a threat to public health. Pharmaco-epidemiology
studies \cite{becker2007hospitalisations} and recent National Health
Statistics Report publications \cite{hall_NHS,Nisha_NHS} indicate
that each year an estimated 195,000 hospitalizations and 74,000 emergency
room visits are the result of DDI in the United States alone \cite{percha_altman}.
DDI has been implicated in nearly 3\% of all hospital admissions \cite{jankel1993epidemiology}
and 4.8\% of admissions among the elderly \cite{becker2007hospitalisations}
and is a common consequence of medical error, representing 3\% to
5\% of all inpatient medication errors \cite{leape1995systems}. With
increasing rates of polypharmacy, which refers to the use of multiple
medications or more medications than are clinically indicated \cite{hajjar2007polypharmacy},
the incidence of DDI will likely increase in the coming years.

Researchers link molecular mechanisms underlying DDI to their clinical
consequences through three types of studies: \emph{in vitro}, \emph{in
vivo}, and clinical \cite{Boyce_part1,Boyce_part2,hennesy2012}. \emph{In
vitro} pharmacology experiments use intact cells (e.g. hepatocytes),
microsomal protein fractions, or recombinant systems to investigate
molecular interaction mechanisms within the cell (i.e. metabolic,
transport- or target-based). \emph{In vivo} studies evaluate whether
such interactions impact drug exposure in humans. Finally, clinical
studies use a population-based approach and large electronic medical
record databases to investigate the contribution of DDI to drug efficacy
and ADR. 

Automated biomedical literature mining (BLM) methods offer a promising
approach for uncovering evidence of possible DDI in published literature
and clinical databases \cite{tatonetti2011detecting}. BLM is a biomedical
informatics methodology that holds the promise of tapping into the
biomedical collective knowledge \cite{abi2008uncovering} by extracting
information from large-scale literature repositories and by integrating
information scattered across various domain-specific databases and
ontologies \cite{shatkay2003mining,jensen2006literature,cohen2008getting}.
It has been used for knowledge discovery in many biomedical domains,
including extraction of protein-protein interactions \cite{leitner2010febs,krallinger2011protein},
protein structure prediction \cite{rechtsteiner2010use}, identification
of genomic locations associated with cancer \cite{mcdonald2004entity},
and mining drug targets \cite{el2008mining}. In the domain of DDI,
putative interactions uncovered by BLM can serve as targets for subsequent
investigation by \emph{in vitro }pharmacological methods as well as
\emph{in vivo} and clinical studies\cite{tatonetti2011detecting}.

BLM has previously been used for DDI information extraction \cite{segura2010resolving,percha2012discovery,Duke_Li_2012,Segura_BMC,segura_11,wu2013integrated},
as overviewed by the literature on recent DDI challenges \cite{segura20111st,segura2014lessons,herrero2013ddi}
and \emph{Pacific Symposium on Biocomputing} sessions \cite{gonzalez2013text,gonzalez2012text}.
However, much remains to be done in automatic extraction of \emph{experimental
evidence of DDI} from text. Importantly, experimental evidence of
DDI is reported differently for the different types of studies described
above. For instance, \emph{in vivo} pharmacokinetic experiments report
parameters such as the \textquoteleft area under the concentration-time
curve\textquoteright , while clinical studies may instead report population-level
statistics of adverse drug reactions. It is important for BLM pipelines
to be able to identify these different kinds of evidence independently.

To address this situation, we demonstrate the use BLM for reliable
extraction of \emph{pharmacokinetic} \emph{evidence} for DDI from
reports of \emph{in vitro} and \emph{in vivo} experiments. Pharmacokinetic
experimental evidence refers to measures of pharmacokinetic parameters
such as the inhibition constant (Ki), the 50\% inhibitory concentration
(IC50), and the area under the plasma concentration-time curve (AUCR).
Such evidence is particularly important in identifying or dismissing
causal mechanisms behind DDIs and in providing support for putative
DDIs extracted from mining patient records, where biases and confounds
in reporting often give rise to non-causal correlations \cite{tatonetti2012data}.
In order to pursue the goal of using BLM to uncover pharmacokinetic
DDI evidence, a collaboration was developed between Rocha's lab, working
on literature mining, and Li's lab, working on\emph{ }pharmacokinetics.
Though this work is focused on pharmacokinetic\emph{ }evidence, in
subsequent studies we will approach other types of DDI evidence (e.g.
clinical evidence).

Our approach is different from previous BLM approaches to DDI information
extraction \cite{segura2010resolving,percha2012discovery,Duke_Li_2012,Segura_BMC,segura_11,wu2013integrated,segura2014lessons}
because our ultimate goal is not to identify interacting drugs themselves
but rather abstracts and sentences containing a \emph{specific type}
of \emph{evidence of drug interaction}.\textbf{ }Existing DDI-extraction
methods and corpora --- including those evaluated under the DDI Extraction
challenges \cite{segura20111st,DDI11,herrero2013ddi,DDI13,segura2014lessons}
--- are not well suited for this task because they do not attempt
to extract \emph{experimental evidence} of drug interactions, nor
specifically label distinct kinds of evidence. For instance, the DDI
Extraction challenge \textquoteleft 11 \cite{DDI11} used a corpus
of several hundred documents from DrugBank \cite{wishart2008drugbank},
but interacting drug pairs were annotated without regard for the presence
of experimental evidence. More recently, the DDI Extraction challenge
\textquoteleft 13 \cite{DDI13} provided a corpus annotated with pharmacokinetic
and pharmacodynamic interactions \cite{herrero2013ddi}, but the goal
of the text mining task was the extraction and classification of interacting
drug pairs, not the extraction of the experimental evidence of interactions.
Other related work has used DrugBank data for large-scale extraction
of drug-gene and drug-drug relationships \cite{Tari_10,percha2012discovery},
and for predicting DDI using a drug-drug network based on phenotypic,
therapeutic, chemical, and genomic feature similarity \cite{cheng2014machine},
but neither study aimed to identify or extract specific kinds experimental
evidence of DDI.

We have previously shown that BLM can be used for automatic extraction
of numerical pharmacokinetics (PK) parameters from the literature
\cite{wang2009literature}. However, that work was not oriented specifically
toward the extraction of evidence of DDI. Recently, we reported high
performance in a preliminary work on automatically classifying PubMed
abstracts that contain pharmacokinetic evidence of DDI \cite{kolchinsky2012evaluation}
(details below). Because identifying relevant abstracts is only a
first step in the process of extracting pharmacokinetic evidence of
DDI, in this work we consider both the problem of identifying abstracts
containing pharmacokinetic evidence of DDI and that of extracting
from abstracts sentences that contain this specific kind of evidence.
In addition to evidence sentence extraction, we also provide a new
assessment of abstract classification using an updated version of
a separately published corpus \cite{wu2013integrated}, leading to
substantially better classification performance than reported in our
preliminary study \cite{kolchinsky2012evaluation}. The updated corpus
is described below and is publicly available. Finally, we provide
a new comparison of classifiers, a new evaluation methodology using
permutation-based significance tests and Principal Component Analysis
(PCA) \cite{wall_SVD} of feature weights, and a detailed study of
the benefits of including features derived from PubMed metadata, named
entity recognition tools and specialized dictionaries. 

We created abstract and sentence corpora using annotation criteria
for identifying pharmacokinetic evidence of DDI. We consider positive
(indicating the presence of interactions) \emph{and} negative (indicating
the absence) DDI evidence as relevant (see ``Materials and Methods''
section), since both provide important information about possible
DDI. Because the criteria considered here are different from those
used in previously available DDI corpora, our results are not directly
comparable to other BLM approaches to DDI. Therefore, we pursued a
thorough evaluation of the performance of different types of classifiers,
feature transforms, and normalization techniques. For both abstract
and sentence classification tasks we tested several linear classifiers:
logistic regression, support vector machines (SVM), binomial Naive
Bayes, linear discriminant analysis, and a modification of the Variable
Trigonometric Threshold (VTT) classifier, previously developed by
Rocha's lab and found to perform well on protein-protein interaction
text mining tasks \cite{lourencco2011linear,kolchinsky2010classification,abi2008uncovering}.
As we describe in the results and discussion sections, classifiers
fall into two main classes based on whether or not they take into
account feature covariances. In addition, we compared different feature
transform methods, including normalization techniques such as \textquoteleft Term
Frequency, Inverse Document Frequency\textquoteright{} (TFIDF) and
dimensionality reduction based on Principle Component Analysis (PCA).
We also compared performance when including features generated by
several Named Entity Recognition (NER) tools and specialized dictionaries. 

In the experiments reported, our goal is to measure the quality of
automated methods in identifying pharmacokinetic evidence of DDIs
reported in the literature. More generally, we seek to demonstrate
that literature mining can be successful in automatically extracting
experimental evidence of interactions as part of DDI workflows. We
show that many classifier configurations achieve high performance
on this task, demonstrating the robustness and efficacy of BLM on
extracting pharmacokinetic evidence of DDI.

\section*{Materials and Methods}

\begin{figure}[H]
\centering{}\includegraphics[width=1\textwidth]{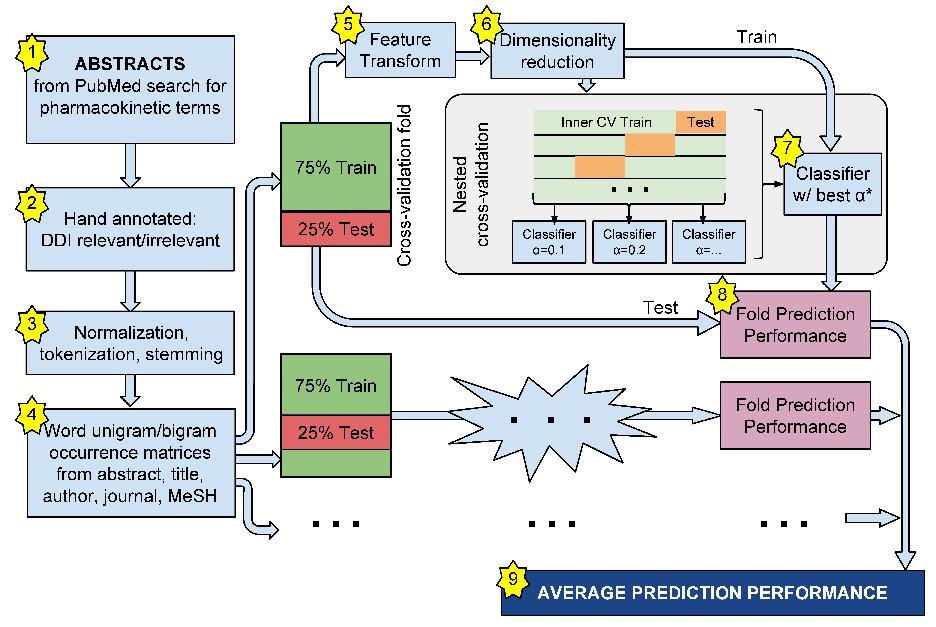}\protect\caption{\label{fig:pipeline}\textbf{Literature mining pipeline}: The basic
steps of the literature mining pipeline include selection of corpus
documents, hand-labeling of ground truth assignments, extraction and
normalization of textual features, and computation of unigram/bigram
occurrences matrices. Cross-validation folds are used to estimate
generalization performance of classifier and feature transform configurations,
while nested (inner) cross-validation folds are used to choose classifier
hyperparameters.}
\end{figure}

The following sections describe the methods used in our literature
mining pipeline. Its basic steps are visually diagrammed in Fig. \ref{fig:pipeline}.
They include the selection of corpus documents, hand-labeling of ground
truth assignments, extraction and normalization of textual features,
and computation of unigram/bigram occurrences matrices. Cross-validation
folds are used to estimate generalization performance of classifier
and feature transform configurations, while nested (inner) cross-validation
folds are used to choose classifier hyperparameters. The software
consisted of custom \texttt{Python} scripts unless otherwise noted.

\subsection*{Abstract Corpus}

For the training corpus, Li's lab selected 1203 pharmacokinetics-related
abstracts by searching PubMed using terms from a previously developed
ontology for PK pharmacokinetic parameters \cite{wang2009literature}.
Therefore, all retrieved articles describe and contain some form of
pharmacokinetic evidence, though not necessarily of DDI. We kept \emph{in
vitro} studies but removed any animal \emph{in vivo} studies. Abstracts
were labeled according to the following criteria: abstracts that reported\emph{
the presence or absence of drug interaction supported by explicit
experimental evidence of pharmacokinetic parameter data} were labeled
as \emph{DDI-relevant} (909 abstracts) while the rest were labeled
as \emph{DDI-irrelevant} (294 abstracts). DDI-relevance was established
regardless of whether the relevant enzymes were presented or not.
Importantly, the concept of DDI-relevance employed here updates the
criteria used in a previous preliminary study \cite{kolchinsky2012evaluation}.
Interactions between a drug and food, fruit, smoking, alcohol, and
natural products are now classified as drug interactions because their
pharmacokinetics studies are designed similarly. For the same reason,
studies dealing with interactions between drug metabolites (instead
of parent compounds) are now also considered relevant, as well as
studies reporting inhibition of induction of a drug on a drug metabolism
enzyme or drug transporter. Classification was done by three graduate
students with M.S. degrees and one postdoctoral annotator; any inter-annotator
conflicts were checked by a Pharm D. and an M.D. scientist with extensive
pharmacological training. The corpus is publicly available as ``Pharmacokinetics
DDI-Relevant Abstracts V0'' in \cite{DDI_Corpus} (see also \cite{wu2013integrated}). 

We extracted textual features from PubMed article title and abstract
text fields as well as the following metadata fields: the author names,
the journal title, the Medical Subject Heading (MeSH) terms, the \textquoteleft registry
number/EC number\textquoteright{} (RN) field, and the \textquoteleft secondary
source\textquoteright{} field (SI) (the latter two fields contain
identification codes for relevant chemical and biological substances).
For each PubMed entry, the content of the above fields was tokenized,
processed by Porter stemming \cite{porter1980algorithm}, and converted
into textual features (unigrams and, in certain runs, bigrams). Strings
of numbers were converted into `\#', short textual features (with
length of less than 2 characters) and infrequent features (that occurred
in less than 2 documents) were omitted. Author names, journal titles,
substance names, and MeSH terms were treated as single textual tokens.

The corpus was represented as binary term-document occurrence matrices.
We evaluated classification performance under two different conditions:
in the first --- referred to as `unigram runs' --- only word unigram
features were used; in the second --- referred to as `bigram runs'
--- word bigram features were used in addition to unigram features.
Bigram runs included a much larger number of parameters (i.e. the
bigram feature coefficients) that needed to be estimated from training
data, which can potentially increase generalization error arising
from increased model complexity \cite{bishop2006pattern}. Testing
the classifiers exclusively with unigram features as well as with
both unigram and bigram features evaluated whether the class information
provided by bigrams outweighed their cost in complexity.

\subsection*{Sentence Corpus}

The evidence sentence task consisted in identifying those sentences
within a PubMed abstract that reported experimental evidence for the
presence or absence of a specific DDI. For this purpose, Li\textquoteright s
group developed a training corpus of 4600 sentences extracted from
428 PubMed abstracts. All abstracts contained (positive or negative)
pharmacokinetic evidence of DDIs. Sentences were manually labeled
as DDI-relevant (1396 sentences) if they \emph{explicitly mentioned
pharmacokinetic evidence for the presence or absence of drug-drug
interactions}, and as DDI-irrelevant (3204 sentences) otherwise. The
same pre-processing and annotation procedures were followed for the
sentence corpus as for the abstract corpus (see section ``Abstract
Corpus''). This corpus is publicly available as ``Deep Annotated
PK Corpus V1'' in \cite{DDI_Corpus} (see also \cite{wu2013integrated}).

\subsection*{Classifiers}

Six different linear classifiers were tested:

1. \emph{VTT}: a simplified, angle-domain version of the \emph{Variable
Trigonometric Threshold} Classifier, previously developed in Rocha's
lab \cite{lourencco2011linear,kolchinsky2010classification,abi2008uncovering}.
Given a document vector \textbf{x}=<\emph{x\textsubscript{1}},...,\emph{x\textsubscript{K}}>
with features (i.e. dimensions) indexed by \emph{i}, the separating
hyperplane is defined as \emph{
\[
\sum_{i}\varphi_{i}x_{i}-\lambda=0
\]
}Here, \emph{\textgreek{l}} is a threshold (bias) and \emph{\textgreek{f}\textsubscript{i}}
is the `angle' of feature \emph{i} in binary class space: 
\[
\varphi_{i}=\arctan\frac{p_{i}}{n_{i}}-\frac{\pi}{4}
\]
where \emph{p\textsubscript{i}} is the probability of occurrence
of feature \emph{i} in relevant-class documents and \emph{n\textsubscript{i}}
is the probability of occurrence of feature \emph{i }in irrelevant-class
documents. The threshold parameter \emph{\textgreek{l}} is chosen
so that a neutral `pseudo-document' defined by \emph{x\textsubscript{i}}=(\emph{p\textsubscript{i}}+\emph{n\textsubscript{i}})/2
falls exactly onto the separating hyperplane.

The full version of VTT, which includes additional parameters to account
for named entity occurrences and which we have previously used in
protein-protein interaction classification, is evaluated in combination
with various NER tools in section ``Impact of NER and PubMed metadata
on abstract classification'' below. VTT performs best on sparse,
positive datasets; for this reason, we do not evaluate it on dense
dimensionality-reduced datasets. Notice that in previous work, we
used a different version of VTT with a cross-validated threshold parameter;
its performance on the tasks was very similar, and is reported in
the Supporting Information as the `VTTcv' classifier (section 1
and 2 in S1 Text).

2. \emph{SVM}: a linear \emph{Support Vector Machine} with a cross-validated
regularization parameter (implemented using the \texttt{sklearn} \cite{scikit-learn}
library's interface to the \texttt{LIBLINEAR} package \cite{fan2008liblinear}).

3. \emph{Logistic regression} classifier with a cross-validated regularization
parameter (also implemented using \texttt{sklearn}'s interface to
\texttt{LIBLINEAR}). 

4. \emph{Naive Bayes} classifier with smoothing provided by a Beta-distributed
prior with a cross-validated concentration parameter.

5. \emph{LDA}: a regularized \emph{Linear Discriminant Analysis} classifier,
following \cite{ye2006efficient}. Singular value decomposition (SVD),
a dimensionality reduction technique, is first used to reduce any
rank-deficiency, after which the covariance matrix is shrunk toward
a diagonal, equal-variance structured estimate. The shrinkage parameter
is determined by cross-validation.

6. \emph{dLDA}: a \textquoteleft diagonal\textquoteright{} LDA, where
only the diagonal entries of the covariance matrix are estimated and
the off-diagonal entries are set to 0. A cross-validated parameter
determines shrinkage toward a diagonal, equal-variance estimate. This
classifier can offer a more robust estimate of feature variances;
it is equivalent to a Naive Bayes classifier with Gaussian features
\cite{bickel2004some}.

Generally, linear classifiers fall into one of two types. Classifiers
of the first type --- sometimes called `naive' in the literature,
which in our case include VTT, dLDA, and Naive Bayes --- learn feature
weights without considering feature covariances. While covariance
information can be useful for distinguishing classes, naive classifiers
often perform well with small amounts of training data, when covariances
are difficult to estimate accurately. Classifiers of the second type
--- which we refer to as `non-naive', and which in our case included
SVM, LDA, and Logistic Regression --- do consider feature covariances
(often in combination with regularization techniques to smooth covariance
estimates) and can achieve superior performance when provided with
sufficient training data.

\subsection*{Feature Transforms}

For both unigram and bigram runs, we evaluated classification performance
on several transforms of the document matrices:

1. No transform: raw binary occurrence matrices (see section ``Abstract
Corpus'').

2. IDF: occurrences of feature \emph{i} were transformed to its Inverse
Document Frequency (IDF) value: $\textrm{idf}\left(i\right)=\log\frac{N}{c_{i}+1}$,
where \emph{c\textsubscript{i}} is the total number of occurrences
of feature \emph{i} among all documents. This reduced the influence
of common features on classification.

3. TFIDF: the Term Frequency, Inverse Document Frequency transform
(TFIDF); same as above, but subsequently divided by the total number
of features that occur in each document. This reduced the impact of
document size differences.

4. Normalization: the non-transformed, IDF, and TFIDF document matrices
underwent a length-normalization transform, where each document vector
was inversely scaled by its L2 norm. L2 normalization has been argued
to be important for good SVM performance \cite{leopold2002text}.

5. PCA: The above matrices were run through a Principal Component
Analysis (PCA) dimensionality reduction step. Projections onto the
first 100, 200, 400, 600, 800, and 1000 components were tested.

Feature transforms can improve classification performance by making
the surfaces that separate documents in different classes more linear
and by decreasing the weight of non-discriminating features. PCA,
on the other hand, reduces the number of parameters that need to be
estimated from training data. If class membership information is contained
in the subspace spanned by the largest principal components, then
this kind of dimensionality reduction can improve generalization performance
by reducing noise and model complexity.

\subsection*{Performance evaluation}

The abstract and sentence corpora described above were used both for
training classifiers and for estimating generalization performance
on out-of-sample documents. In order to estimate out-of-sample performance,
we used the following cross-validation procedure for each possible
classifier and feature transform:

1. Each corpus was randomly partitioned into 4 document folds (75\%-25\%
splits). This was repeated 4 times, yielding 16 \emph{outer folds}.
All classifiers and transforms were evaluated using the same partitions. 

2. For each fold, the 75\% split was treated as the `training' split
and the 25\% split was treated as the `testing split'. If a feature
transform was used, it was applied to both splits but was computed
using statistics (such as IDF or principal components) from the training
split. Finally, classifiers were trained on the training split and
evaluated based on their prediction performance on the testing split.

3. Measures of classification performance (see below) on the testing
split were collected. The 16 sets of performance measures were averaged
to produce an estimate of generalization performance.

Because training and testing documents are always separated, for each
cross-validation fold the above procedure is equivalent to calculating
performance on an independent testing corpus.

Except for VTT, the classifiers listed in section ``Classifiers''
used cross-validated regularization parameters. These parameters were
not chosen using cross-validation on the outer folds because this
would lead to a biased estimate of out-of-sample performance. Instead,
regularization parameters were chosen using nested cross-validation
within each of the 75\% blocks of the above outer folds:

1. The 75\%-block was itself partitioned into 4 folds (75\%-25\% splits
of the outer 75\% block). This is repeated 4 times, producing a total
of 16 \emph{inner folds} for each outer fold training split.

2. Over a range of values of the cross-validated parameter, the procedure
described in step 2 above was used, but now applied to the 75\%/25\%
splits of each \emph{inner fold. }Mean performance on inner fold testing
splits were measured using the Matthews Correlation Coefficient \cite{matthews1975comparison}
(MCC), which is particularly well-suited for the unbalanced scenarios
of our corpora\cite{Baldi2000a}. 

3. The parameter value giving the highest mean MCC was chosen as the
regularization parameter value for training the classifier in the
outer fold.

We evaluated the performance of the classifiers using three different
measures: the balanced F1 score (the harmonic mean of precision and
recall), the iAUC or `area under the interpolated precision/recall
curve' \cite{davis2006relationship}, and the MCC. In addition, we
computed and reported the rank product of these three measures (RP3)
as a single inclusive metric of classification performance. The RP3
measure provides a well-rounded assessment of classifier performance,
as it combines the ranking of the different individual measures \cite{lourencco2011linear,abi2008uncovering}. 

For displaying results, we focus primarily on the iAUC measure (in
cases where only plots of iAUC performance are provided, F1 and MCC
plots are found in the Supporting Information, S1 Text). iAUC does
not depend on predicted class assignments but rather on the ranking
of test set documents according to classifier confidence scores from
most relevant to most irrelevant. iAUC offers three major advantages
as a measure of classification performance. First, it provides a more
comprehensive measure of classifier performance because it evaluates
the entire ranking of documents, as opposed to just class assignments.
Second, iAUC is less sensitive to variation driven by random-sampling
differences in the training corpus, which may lead to fluctuations
in the class assignments of low confidence documents and, correspondingly,
high variability in measures such as F1 and MCC. Finally, it is more
relevant in a frequently-encountered situation where a human practitioner
uses a BLM pipeline to retrieve only the most relevant documents (which
should have high positive-class confidence scores) or to identify
likely-to-be-misclassified documents (which should have low confidence
scores).

Both the abstract and sentence classification tasks are characterized
by imbalanced datasets, with more relevant-class abstracts and more
irrelevant-class sentences respectively. For simplicity, and because
we are primarily concerned how ranking performance (as measured by
iAUC) changes between different machine learning configurations on
the same dataset, we do not perform resampling or re-weighting of
training items. We also report MCC values, a measure which is known
to be stable in the face of unbalanced classes \cite{Baldi2000a}.

The performance of a classifier and feature transform configuration
varies both due to random sampling of folds and due to the inherent
performance bias of the configuration over the entire distribution
of folds. Since we are only interested in the latter, observed performance
differences between pairs of configurations were tested for statistical
significance using a non-parametric paired-sample permutation test.
First, the assignments of performance scores for each of the 16 outer
folds were permuted between the two classifier/transform configurations
under consideration. For each of the 2\textsuperscript{16} possible
permutations, the difference in across-fold mean performance was calculated;
this formed the distribution of performance differences under the
null hypothesis that the two configurations have equal performance.
Finally, the \emph{p}-value was computed as the probability (one-
or two-tailed, as indicated) of observing a difference under the null
hypothesis distribution equal to or greater than the actual difference.

\section*{Results}

\subsection*{Abstract classification performance}

\begin{figure}[H]
\begin{centering}
\includegraphics[width=1\textwidth]{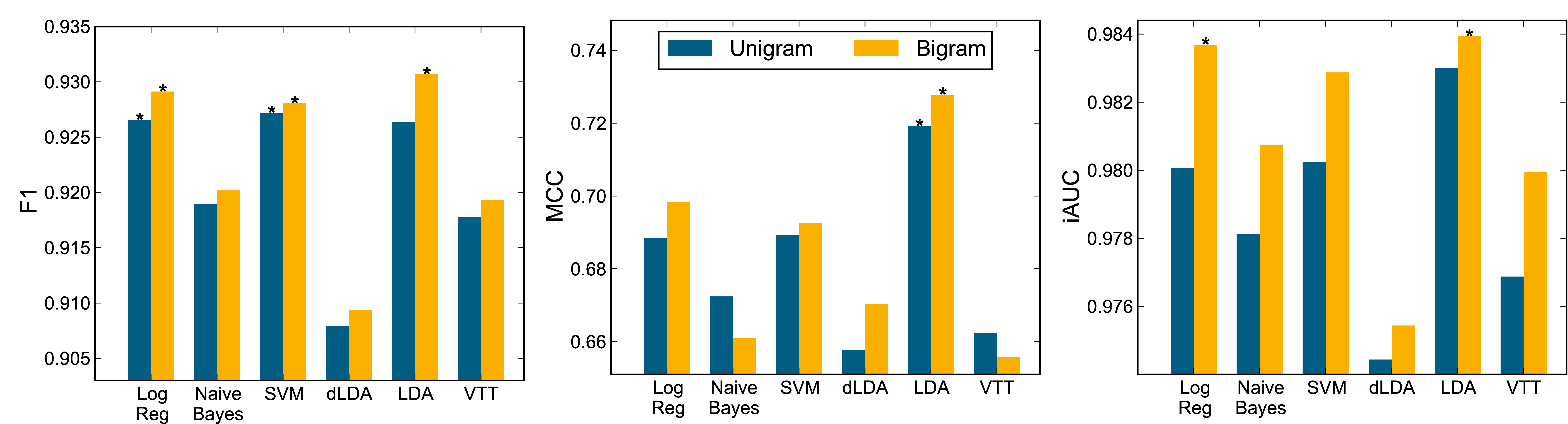}
\par\end{centering}

\protect\caption{\label{fig:performance}\textbf{Classification performance on abstracts}:
Performance for both unigram and bigram runs on non-transformed features.
Left: F1 measure. Middle: MCC measure. Right: iAUC measure. The best
classifier configuration, and configurations not significantly different
(\emph{p}>0.05, one-tailed test) from it, marked with asterisk `{*}'.}

\end{figure}

Fig. \ref{fig:performance} shows classifier performance on the abstract
task for the unigram and bigram runs with no feature transform applied.
The best classifier configuration, as well as those configurations
not significantly different from the best (\emph{p}>0.05, one-tailed
test), are marked with an asterisk. In addition, the performance results,
ranks, and the rank-product (RP3) measure are reported in Table \ref{tab:abstable}.
The best classifier achieves F1\ensuremath{\approx}0.93, iAUC\ensuremath{\approx}0.98,
MCC\ensuremath{\approx}0.73, which constitutes a substantial and significant
improvement over our previous preliminary results reported in \cite{kolchinsky2012evaluation},
where we had reached F1\ensuremath{\approx}0.8, iAUC\ensuremath{\approx}0.88,
MCC\ensuremath{\approx}0.6 (notice that these performance values would
be well below the lowest reported levels in Fig. \ref{fig:performance}).
This demonstrates that the corpus used in this work --- which is more
carefully curated and now also considers interactions between drugs
and food, fruit, smoking, alcohol, and natural products to be relevant
(details in ``Materials and Methods Section'') --- improves the
classification of abstracts with pharmacokinetic evidence of DDI.
The levels of performance achieved are excellent when compared to
similar abstract classification tasks in other biomedical domains.
For instance, in the BioCreative Challenge III, considered one of
the premier forums for assessment of text mining methods, the best
classifiers of abstracts with Protein-Protein Interaction yielded
performances of F1\ensuremath{\approx}0.61, iAUC\ensuremath{\approx}0.68,
MCC\ensuremath{\approx}0.55 \cite{krallinger2011protein}. Naturally,
our results are not directly comparable to results obtained on different
corpora and on a different problem; rather, these numbers provide
guidance on what is typically considered good results in biomedical
article classification.

\begin{table}[H]
\begin{centering}
\textsf{\footnotesize{}\begin{tabular}{l l l l l r}
\textbf{Classifier} & \textbf{Type} & \textbf{F1} & \textbf{MCC} & \textbf{iAUC} & \textbf{RP3} \\
\hline \hline
LDA & Bigram & .931 (1) & .728 (1) & .984 (1) & 1 \\
\hline 
Log Reg & Bigram & .929 (2) & .698 (3) & .984 (1) & 6 \\
\hline 
SVM & Bigram & .928 (3) & .693 (4) & .983 (3) & 36 \\
\hline 
LDA & Unigram & .926 (6) & .719 (2) & .983 (3) & 36 \\
\hline 
Log Reg & Unigram & .927 (4) & .689 (5) & .980 (6) & 120 \\
\hline 
SVM & Unigram & .927 (4) & .689 (5) & .980 (6) & 120 \\
\hline 
Naive Bayes & Bigram & .920 (7) & .661 (10) & .981 (5) & 350 \\
\hline 
Naive Bayes & Unigram & .919 (8) & .672 (7) & .978 (9) & 504 \\
\hline 
VTT & Bigram & .919 (8) & .656 (12) & .980 (6) & 576 \\
\hline 
VTT & Unigram & .918 (10) & .662 (9) & .977 (10) & 900 \\
\hline 
dLDA & Bigram & .909 (11) & .670 (8) & .975 (11) & 968 \\
\hline 
dLDA & Unigram & .908 (12) & .658 (11) & .974 (12) & 1584 \\
\hline 
\end{tabular}}
\par\end{centering}{\footnotesize \par}

\smallskip{}
\protect\caption{\label{tab:abstable}\textbf{Classification performance on abstracts}:
Performance for both unigram and bigram runs on non-transformed features
according to F1, MCC, and iAUC performance measures. The rank of the
classifiers according to each measure is reported in parenthesis in
the respective column. Classifiers are ordered according to the rank
product (RP3) of the three measures (last column).}
\end{table}

For each classifier, the inclusion of bigram features improved performance
according to the RP3 measure. The best classifier according to all
measures was LDA using bigrams. The performance of this classifier
was significantly better than all others for the MCC, but not significantly
better that Logistic Regression according to iAUC, and not significantly
better that Logistic Regression and SVM according to F1 score. According
to RP3, these three classifiers using bigrams yield the best performance.
Naive Bayes, VTT, and dLDA --- classifiers that make a `naive' independence
assumption about features (see ``Materials and Methods'' section)
--- performed below the top three. However, the performance levels
they achieved are still quite high, which indicates that such simple
classifiers are also capable of classifying documents with pharmacokinetic
DDI evidence in our corpus. The in-house VTT classifier is the only
classifier among these that does not use cross-validated parameters;
when used with NER features and cross-validated parameters (the configuration
for which it was originally designed \cite{lourencco2011linear,kolchinsky2010classification,abi2008uncovering}),
its performance improved (see below).

\subsubsection*{Feature Transforms and Dimensionality Reduction}

The different feature transforms and PCA-based dimensionality reductions
(section ``Materials and Methods'') significantly improved performance
for several classifiers, though they could not beat the performance
of the best non-transformed classifier. Details are provided in Supporting
Information (section 1.1 in S1 Text). To summarize, according to most
measures only dLDA and SVM improved performance significantly with
either an IDF or TFIDF transform plus L2 normalization and dimensionality
reduction (top \emph{n} principal components). For instance, the best
iAUC for SVM (0.984) occurs with a dimensionality reduction to the
top 800 principal components and no feature transform; this is a significant
improvement over the no-transform, no dimensionality reduction SVM
classifier reported in Table \ref{tab:abstable} and Fig. \ref{fig:performance},
but not a significant improvement over the overall best classifiers
reported there (LDA and Logistic regression). The dLDA classifier
significantly improves its iAUC performance with almost all feature
transform and dimensionality reduction combinations, but not above
that of the top performing classifiers. We conclude that feature transforms
and dimensionality reduction does not lead to the best classification
performance on the abstract task.

\subsubsection*{Pharmacokinetics DDI Features in abstract classification}

We looked at which textual features play the largest role in the abstract
classification task. A linear classifier separates document classes
with a hyperplane defined by a set of feature coefficients. The impact
of a feature on classification is quantified by the sign and amplitude
of its hyperplane coefficient. A feature with a large positive coefficient
contributes strongly to a document's propensity to be classified as
relevant, while a feature with a large negative coefficient contributes
strongly to a document's propensity to be classified as irrelevant.
In Table \ref{tab:abstop10features}, we show the top 20 most distinctive
features of the relevant and irrelevant classes in the abstract task,
as chosen in the bigrams runs by the LDA classifier (left) and Logistic
Regression classifier (right), the two top-performing classifiers
in this task according to the RP3 measure (see Table \ref{tab:abstable}).
Notice that textual features are stemmed.

\begin{table}[H]
\begin{minipage}[t]{0.5\columnwidth}%
\begin{center}
\textbf{LDA (Bigram)}\textsf{\footnotesize{}}\\
\textsf{\footnotesize{}\begin{tabular}{c c}
\textbf{Relevant} & \textbf{Irrelevant} \\
\hline \hline
\emph{MeSH:Drug Interactions} & area \\
\hline 
interact & rate \\
\hline 
inhibit & differ \\
\hline 
interact between & polymorph \\
\hline 
oral & activ \\
\hline 
day & genotyp \\
\hline 
decreas & higher \\
\hline 
receiv & patient with \\
\hline 
mg & conclus \\
\hline 
increas & to the \\
\hline 
auc & \emph{Substance:Hydrocarb. Hydroxylas.} \\
\hline 
inhibitor & that the \\
\hline 
\emph{MeSH:Male} & lower \\
\hline 
treatment & patient \\
\hline 
chang & allel \\
\hline 
increas the & among \\
\hline 
dure & extens \\
\hline 
on the & \emph{MeSH:Female} \\
\hline 
alon & of the \\
\hline 
combin & analysi \\
\hline 
\end{tabular}}
\par\end{center}{\footnotesize \par}%
\end{minipage}%
\begin{minipage}[t]{0.5\columnwidth}%
\begin{center}
\textbf{Logistic Regression (Bigram)}\textsf{\footnotesize{}}\\
\textsf{\footnotesize{}\begin{tabular}{c c}
\textbf{Relevant} & \textbf{Irrelevant} \\
\hline \hline
\emph{MeSH:Drug Interactions} & area \\
\hline 
interact & rate \\
\hline 
inhibit & differ \\
\hline 
interact between & \emph{MeSH:Reference Values} \\
\hline 
oral & activ \\
\hline 
decreas & that the \\
\hline 
mg & conclus \\
\hline 
\emph{Substance:Enzyme Inhibitors} & clearanc of \\
\hline 
receiv & higher \\
\hline 
auc & patient \\
\hline 
determin & \emph{MeSH:Female} \\
\hline 
treatment & patient with \\
\hline 
day & \emph{MeSH:Injections, Intravenous} \\
\hline 
inhibitor & polymorph \\
\hline 
chang & genotyp \\
\hline 
dure & \emph{MeSH:Phenotype} \\
\hline 
alon & among \\
\hline 
\emph{MeSH:Male} & healthi subject \\
\hline 
administ & \emph{MeSH:Half-life} \\
\hline 
chang in & lower \\
\hline 
\end{tabular}}
\par\end{center}{\footnotesize \par}%
\end{minipage}

\smallskip{}
\protect\caption{\label{tab:abstop10features}\textbf{Top 20 relevant and irrelevant
abstract features}: The stemmed textual features most discriminative
of relevant and irrelevant classes on the abstract task, as chosen
by two of the top-performing classifiers according to the RP3 measure:
LDA with bigrams (left) and Logistic Regression with bigrams (right).}
\end{table}

Some of the most relevant features come from MeSH term metadata (such
as the MeSH term \emph{Drug interactions}) and terms that explicitly
indicate interactions (\textquoteleft interact\textquoteright , \textquoteleft inhibit\textquoteright ,
\textquoteleft interact between\textquoteright , \textquoteleft decreas\textquoteright ,
\textquoteleft increas\textquoteright ). Other relevant terms deal
with administration protocols and study design (\textquoteleft oral\textquoteright ,
`day', `receiv', `mg', `treatment\textquoteright , \textquoteleft alon\textquoteright ,
\textquoteleft combin\textquoteright ). Some of the irrelevant features
concern genetics terminology (`allel', `genotyp', `polymorph',
and MeSH term \emph{Phenotype}), indicating that the irrelevant class
was enriched with genetics or pharmacogenetics vocabulary. Several
generic biomedical terms (such as `patient', `healthi subject',
`higher') terms are also highly irrelevant. In addition, highly
irrelevant features also contain some non-DDI-specific pharmacokinetic
terms (for example, \textquoteleft area\textquoteright , \textquoteleft rate\textquoteright ,
`clearance of\textquoteright ), which is not surprising given that
both relevant and irrelevant articles were drawn from pharmacokinetics-related
literature. One surprising result is the observation that while the
MeSH term \emph{Mal}e is one of the top relevant features, the MeSH
term \emph{Female} is one of the top irrelevant features. We have
no explanation for the cause of this gender imbalance since the corpus
was built from automatic searches to PubMed without any gender-specific
query terms. 

Further analysis of highly relevant and irrelevant features across
all classifiers and feature transforms was performed and reported
in the Supporting Information (section 1.2 in S1 Text). We quantified
and plotted the contribution of standardized coefficients \cite{agresti2007introduction}
of different features and show the most positively and negatively
loaded features for different classifier and transform configurations.
Top textual features obtained from all classifiers include additional
terms falling under the categories described above, with features
derived from PubMed metadata (MeSH, chemical substances) also appearing
among both the most relevant and irrelevant sets. Other relevant MeSH
terms, besides \emph{Drug Interactions}, include \emph{Cimetidine/pharmacology},
\emph{Cross-Over Studies}, \emph{Enzyme Inhibitors/PK}, \emph{Kinetics},
and \emph{Proton Pump Inhibitors}. Additionally, a PubMed author entry
corresponding to a prominent researcher in the pharmacokinetics DDI
field (\textquoteleft PJ Neuvonen\textquoteright ) appears as highly
relevant, as well as three substances from the RN field (see also
section ``Impact of NER and PubMed metadata on abstract classification''
below): \emph{Cimetidine}, \emph{Enzyme Inhibitors}, and \emph{Proton
Pump Inhibitors}. For the irrelevant set, additional MeSH (\emph{Anti-Ulcer
Agents/adm\&dos}; \emph{Injections, Intravenous}; \emph{Phenotype};
\emph{Protein Binding}; \emph{Reference Values}) and Substance terms
also appear (\emph{Anti-ulcer agents}; \emph{Hydrocarb. Hydroxylas}).
The Supporting Information S1 Text contains details of the analysis
and lists of features. It also shows the results of a Principal Component
Analysis of feature weight coefficients chosen by different classifiers.

\subsubsection*{Impact of NER and PubMed metadata on abstract classification}

We have previously demonstrated improved classification performance
on protein-protein interaction BLM tasks by supplementing textual
features (such as the word unigram and bigram occurrences) with features
built using \emph{Named Entity Recognition} (NER) and domain-specific
\emph{dictionary} tools \cite{lourencco2011linear,kolchinsky2010classification,abi2008uncovering}.
To test if similar techniques are useful in the DDI domain, we counted
mentions of named biochemical species (e.g. proteins, compounds and
drugs) and concepts (e.g. pharmacokinetic terms) in each document
and then included these counts as document features in addition to
the bigram and unigram textual features. Counts were extracted using
biomedical-specific NER extraction tools and dictionaries, with dictionary
matches identified by internally-developed software. A preliminary
study of the impact of NER/Dictionary features was reported in \cite{kolchinsky2012evaluation}
using a previous less-refined DDI corpus. Here, in addition to using
the more fine-tuned corpus (see ``Methods and Data'' section), we
study the impact of PubMed metadata features on classification performance.
We also provide a new comprehensive analysis of the performance impact
of including features from several publicly-available NER and metadata
resources:
\begin{itemize}
\item OSCAR4 \cite{jessop2011oscar4}: NER tool for chemical species, reaction
names, enzymes, chemical prefixes and adjectives.
\item ABNER \cite{settles2005abner}: NER tool for genes, proteins, cell
lines and cell types.
\item BICEPP \cite{lin2011bicepp}: NER tool for clinical characteristics
associated with drugs.
\item \emph{DrugBank} database \cite{wishart2006drugbank}: a dictionary
list of drug names
\item \emph{Dictionaries} provided by Li's lab. \emph{i-CYPS}: cytochrome
P450 {[}CYP{]} protein names, a group of enzymes centrally involved
in drug metabolism; \emph{i-PkParams}: terms relevant to pharmacokinetic
parameters and studies; \emph{i-Transporters}: proteins involved in
transport; \emph{i-Drugs}: Food and Drug Administration's drug names.
The dictionaries are available for download from \cite{DDI_Corpus}.
\end{itemize}
For each of these NER tools and dictionaries, we counted the number
of occurrences of any of its entities/entries in a given abstract.
These counts were treated as any other feature for SVM, Logistic Regression,
diagonal LDA, and LDA classifiers. Naive Bayes was omitted since NER
count features are non-binary. VTT incorporates NER features via a
modified separating hyperplane equation:
\[
\sum_{i}\varphi_{i}x_{i}-\sum_{j}\frac{\beta_{j}-c_{j}}{\beta_{j}}-\lambda=0
\]
where \emph{x\textsubscript{i}} represents the occurrence of textual
feature \emph{i}, \emph{\textgreek{f}\textsubscript{i}} and \emph{\textgreek{l}}
are textual feature and bias parameters as described in section ``Classifiers'',
\emph{c\textsubscript{j}} is the count of NER/Dictionary feature
from resource \emph{j}, and \emph{\textgreek{b}\textsubscript{j}}
is a weight for resource \emph{j}, which is chosen by cross-validation.

\begin{figure}[H]
\includegraphics[width=1\textwidth]{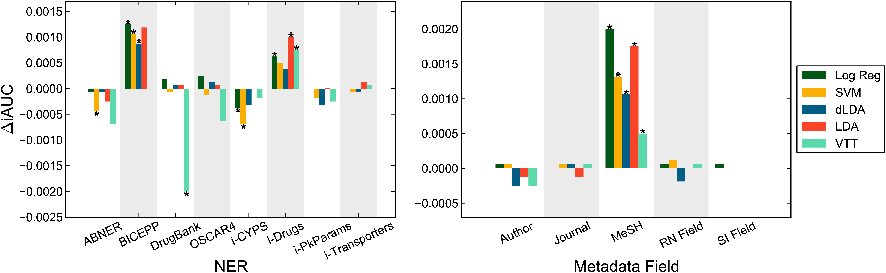}\protect\caption{\label{fig:NER-performance}\textbf{Performance impact of abstract
NER and metadata features}: Left: Relative changes in iAUC scores
on non-transformed bigram runs in combination with different NER/Dictionary
features. Significant changes (\emph{p}<0.05, two-tailed test) in
performance over the respective classifiers without NER features are
indicated with asterisk `{*}'. Right: Relative changes in iAUC when
features from a given PubMed metadata field are included versus omitted
(while including features from the other 4 metadata fields). Significant
changes (\emph{p}<0.05, two-tailed test) in performance are indicated
with asterisk `{*}'.}
\end{figure}

In Fig. \ref{fig:NER-performance} (left), we plot the relative iAUC
changes over the respective classifiers without NER/Dictionary count
features (results for MCC and F1 in Supporting Information; section
1.3 in S1 Text). Significant performance changes are indicated with
an asterisk (\emph{p}<0.05, two-tailed test). Some NER/Dictionary
features improved performance significantly for several classifiers.
However, the inclusion of two dictionary features (\emph{DrugBank},
and \emph{i-CYPS}) actually decreased performance significantly for
several classifiers, suggesting that these features contain little
class information and instead contribute to over-fitting. Table \ref{tab:abstable_NER}
lists performance for configurations in which NER and dictionary features
gave a significant performance increase for at least one of the three
measures (F1, MCC, or iAUC), along with best classifier performance
using only textual features (bigram runs). The BICEPP tool consistently
yielded the best improvement for every classifier tested, followed
by the \emph{i-Drugs} dictionary. The OSCAR4 tool also significantly
improved the performance of the VTT classifier (especially for the
MCC measure as shown in Supporting Information, S1 Text). With the
inclusion of NER and dictionary features, the overall top classifiers
(LDA and Logistic Regression), significantly improved their performance,
now reaching F1\ensuremath{\approx}0.93, MCC\ensuremath{\approx}0.74,
iAUC\ensuremath{\approx}0.99. Among the set of naive classifiers,
VTT improved performance significantly with the inclusion of NER features,
ranking above the other naive classifiers according to the RP3 measure.

\begin{table}[H]
\begin{centering}
\textsf{\footnotesize{}\begin{tabular}{l l l l l r}
\textbf{Classifier} & \textbf{NER} & \textbf{F1} & \textbf{MCC} & \textbf{iAUC} & \textbf{RP3} \\
\hline \hline
LDA & BICEPP & .933 (2) & .737 (1) & .985 (1) & 2 \\
\hline
LDA & \emph{i-Drugs} & .934 (1) & .736 (2) & .985 (1) & 2 \\
\hline
Log Reg & BICEPP & .933 (2) & .714 (3) & .985 (1) & 8 \\
\hline
Log Reg & \emph{i-Drugs} & .930 (6) & .700 (6) & .985 (1) & 36 \\
\hline
LDA & $-$ & .931 (5) & .728 (3) & .984 (5) & 75 \\
\hline
SVM & BICEPP & .932 (4) & .710 (5) & .984 (5) & 100 \\
\hline
Log Reg & $-$ & .929 (8) & .698 (7) & .984 (5) & 280 \\
\hline
SVM & \emph{i-Drugs} & .930 (6) & .687 (10) & .984 (5) & 300 \\
\hline
SVM & $-$ & .928 (9) & .693 (8) & .983 (9) & 648 \\
\hline
VTT & BICEPP & .922 (11) & .692 (9) & .980 (12) & 1188 \\
\hline
VTT & OSCAR4 & .923 (10) & .683 (11) & .979 (14) & 1540 \\
\hline
VTT & \emph{i-Drugs} & .920 (12) & .670 (14) & .981 (10) & 1680 \\
\hline
Naive Bayes & $-$ & .920 (12) & .661 (16) & .981 (10) & 1920 \\
\hline
dLDA & BICEPP & .911 (15) & .680 (12) & .976 (15) & 2700 \\
\hline
VTT & $-$ & .919 (14) & .656 (17) & .980 (12) & 2856 \\
\hline
dLDA & \emph{i-Drugs} & .911 (15) & .678 (13) & .975 (16) & 2700 \\
\hline
dLDA & $-$ & .909 (17) & .670 (14) & .975 (16) & 3808 \\
\hline
\end{tabular}}
\par\end{centering}{\footnotesize \par}

\smallskip{}
\protect\caption{\label{tab:abstable_NER}\textbf{Abstract classification performance
using NER features}: Performance of the best classifiers when specific
NER and dictionary features are added; original (bigram runs) classifiers
also listed with no NER features (indicated by -). F1, MCC, and iAUC
performance measures are listed; the rank of the classifiers according
to each measure is reported in parenthesis in the respective column.
Classifiers are ordered according to the rank product (RP3) of the
three measures (last column).}
\end{table}

As mentioned, word unigram and bigram features were extracted not
only from article abstracts and titles, but also from five PubMed
metadata fields: author names, journal titles, MeSH terms, and two
fields referring to standardized substance names: the \textquoteleft registry
number/EC number\textquoteright{} {[}RN{]} field and the \textquoteleft secondary
source\textquoteright{} field {[}SI{]}. In fact, some PubMed metadata
features were among those most distinguishing of relevant and irrelevant
abstracts (for greater detail, see Table \ref{tab:abstop10features}
and section ``Pharmacokinetics DDI Features in abstract classification'',
as well Supporting Information; section 1.2 in S1 Text). We tested
the impact of PubMed metadata fields on abstract classification performance.
In Fig. \ref{fig:NER-performance} (right), we plot the relative iAUC
changes when features from a given PubMed metadata field are included
versus omitted (while including features from the other 4 metadata
fields).\textbf{ }Significant changes (\emph{p}<0.05, two-tailed test)
in performance are indicated with an asterisk; results for MCC and
F1 can be found in Supporting Information (S1 Text). MeSH terms was
the only metadata source whose omission decreased performance significantly.
However, the performance increase of including MeSH data is rather
small. Therefore, the methodology does not require the availability
of human-annotated metadata such as MeSH terms and can still be deployed
on recent articles that have not yet been annotated with MeSH terms.

\subsection*{Evidence sentence extraction performance}

\begin{figure}[h]
\includegraphics[width=1\textwidth]{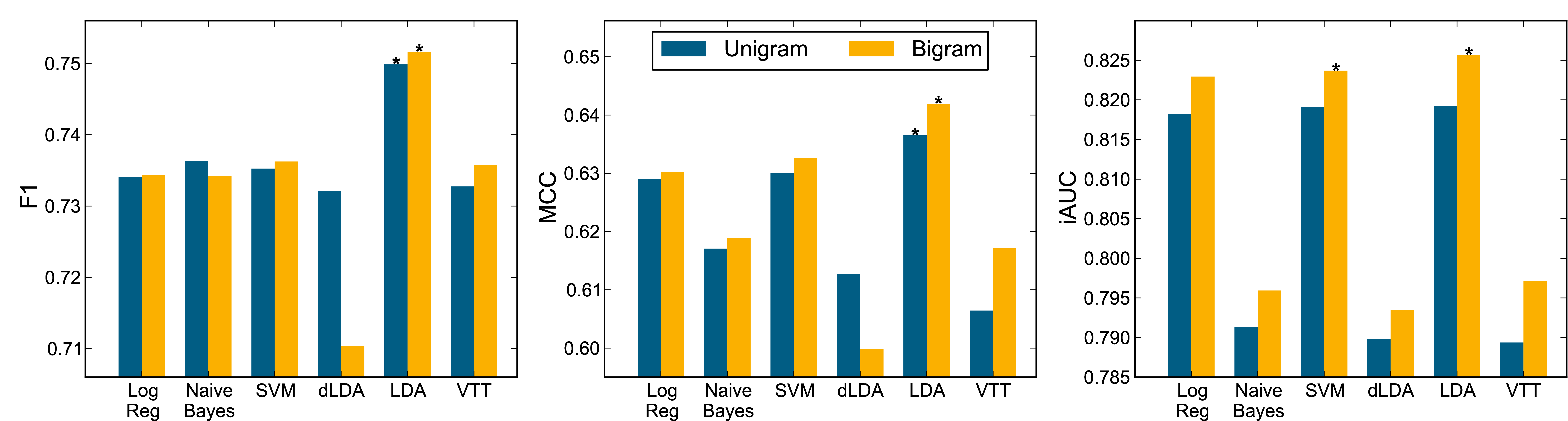}\protect\caption{\label{fig:performance-sentences}\textbf{Sentence classification
performance}: Performance for both unigram and bigram runs on non-transformed
features. Left: F1 measure. Middle: MCC measure. Right: iAUC measure.
The best classifier configuration, and configurations not significantly
different (\emph{p}>0.05, one-tailed test) from it, marked with asterisk
`{*}'.}
\end{figure}

Fig. \ref{fig:performance-sentences} shows classification performance
on the sentence task of the unigram and bigram runs without any feature
transforms applied, according to F1, MCC, and iAUC measures. The best
classifier configuration, as well as those configurations not significantly
different from the best (\emph{p}>0.05, one-tailed test), are marked
with an asterisk. In addition, the numerical results, ranks, and the
rank-product (RP3) measure are reported in Table \ref{tab:senttable}.

\begin{table}[H]
\begin{centering}
\textsf{\footnotesize{}\begin{tabular}{l l l l l r}
\textbf{Classifier} & \textbf{Type} & \textbf{F1} & \textbf{MCC} & \textbf{iAUC} & \textbf{RP3} \\
\hline \hline
LDA & Bigram & .752 (1) & .642 (1) & .826 (1) & 1 \\
\hline 
LDA & Unigram & .750 (2) & .636 (2) & .819 (4) & 16 \\
\hline 
SVM & Bigram & .736 (3) & .633 (3) & .824 (2) & 18 \\
\hline 
Log Reg & Bigram & .734 (7) & .630 (4) & .823 (3) & 84 \\
\hline 
SVM & Unigram & .735 (6) & .630 (4) & .819 (4) & 96 \\
\hline 
VTT & Bigram & .736 (3) & .617 (8) & .797 (7) & 168 \\
\hline 
Naive Bayes & Unigram & .736 (3) & .617 (8) & .791 (10) & 240 \\
\hline 
Log Reg & Unigram & .734 (7) & .629 (6) & .818 (6) & 252 \\
\hline 
Naive Bayes & Bigram & .734 (7) & .619 (7) & .796 (8) & 392 \\
\hline 
dLDA & Unigram & .732 (11) & .613 (10) & .790 (11) & 1210 \\
\hline 
dLDA & Bigram & .710 (12) & .600 (12) & .794 (9) & 1296 \\
\hline 
VTT & Unigram & .733 (10) & .606 (11) & .789 (12) & 1320 \\
\hline 
\end{tabular}}
\par\end{centering}{\footnotesize \par}

\medskip{}
\protect\caption{\label{tab:senttable}\textbf{Sentence classification performance}:
Performance for both unigram and bigram runs on non-transformed features
according to F1, MCC, and iAUC performance measures. The rank of the
classifiers according to each measure is reported in parenthesis in
the respective column. Classifiers are ordered according to the rank
product (RP3) of the three measures (last column).}
\end{table}

As with abstracts, including bigram features tended to improve sentence
classification performance. LDA performed best, having the highest
RP3 and being the best classifier according to the F1 and MCC measures
and one of the two best classifiers (along with SVM) on the iAUC measure.
Generally, the classifiers that performed well on the sentence task
were those that took into account feature covariances: SVM, Logistic
Regression, and LDA. The top classifier (LDA with bigrams) on the
evidence sentence task reached performance of F1\ensuremath{\approx}0.75,
MCC\ensuremath{\approx}0.64, iAUC\ensuremath{\approx}0.83. 

We measured sentence classification performance in combination with
different feature transforms and dimensionality reductions (see section
2.1 in Text S1). In general, the three classifiers that do best on
non-transformed features (SVM, Logistic Regression, and LDA) show
decreased performance with dimensionality reduction according to all
measures, with more extreme dimensionality reduction leading to larger
performance decreases. On the other hand, for dLDA (a `naive' classifier
that treats features as independent), PCA-based dimensionality reduction
--- which uses feature covariances to choose optimal projections ---
led to significant improvements in all measures, with more dimensions
giving better performance. These findings indicate that the pattern
of feature covariance carries important information about class membership
in the sentence task, and that this pattern is distributed across
a large number of dimensions. Generally, the LDA classifier achieved
the best performance according to all three measures. Its baseline
performance according to the iAUC measure was further improved significantly
by an IDF-transform, and --- according to the F1 measure --- by any
transform containing an L2 normalization.

In Table \ref{tab:senttop10features}, we show the top 20 features
most distinctive of the relevant and irrelevant classes in the sentence
task as chosen by the LDA classifier on bigrams (the top performing
classifier according to the RP3 measure; features of other top classifiers
are not shown because they were highly similar). Numerical features
(indicated by `\#.\#') were highly indicative of the relevant class,
along with expressions of quantitative changes (`decreas', `increas')
and interaction (`inhibit', `catalyz', `interact with') as well
as adverbs expressing significance of evidence (`significantli').
Also highly relevant were features referring to the area under the
concentration-time curve (`auc'), which is often employed in pharmacokinetics
to measure differences in drug clearance rates under different experimental
conditions. Names of several drugs (`ketoconazol', `itraconazol',
`quiindin') were relevant in predicting DDI evidence sentences.
These drugs are frequently used probe inhibitors for metabolism enzymes
CYP3A4/5, CYP3A4/5 and CYP2D6 respectively and are routinely used
in drug interaction studies.

\begin{table}[H]
\begin{minipage}[t]{0.5\columnwidth}%
\begin{center}
\textbf{LDA (Bigram)}\textsf{\footnotesize{}}\\
\textsf{\footnotesize{}\begin{tabular}{c c}
\textbf{Relevant} & \textbf{Irrelevant} \\
\hline \hline
inhibit & day \\
\hline 
increas & investig \\
\hline 
\#.\# & determin \\
\hline 
ketoconazol & vitro \\
\hline 
decreas & evalu \\
\hline 
microm & enzym \\
\hline 
rifampin & use \\
\hline 
format & differ \\
\hline 
catalyz & cytochrom p450 \\
\hline 
auc & studi \\
\hline 
significantli & dose \\
\hline 
coadministr & examin \\
\hline 
itraconazol & measur \\
\hline 
quinidin & subject \\
\hline 
clearanc & assess \\
\hline 
reduc & interact \\
\hline 
\#.\#-fold & compar \\
\hline 
show & drug \\
\hline 
co-administr & genotyp \\
\hline 
interact with & cytochrom \\
\hline 
\end{tabular}}
\par\end{center}{\footnotesize \par}%
\end{minipage}

\smallskip{}
\protect\caption{\label{tab:senttop10features}\textbf{Top 20 relevant and irrelevant
sentence features}: The most discriminative features of relevant and
irrelevant classes in the sentence task, as chosen by the top-performing
classifiers according to the RP3 measure: LDA on bigrams.}
\end{table}

Highly irrelevant features refer to more generic pharmacokinetic or
biomedical concepts such as `investig', `dose', `enzym', `studi',
etc. Interestingly, some terms that are highly relevant in the abstract
task are highly irrelevant in the sentence task (e.g., `day'). Notably,
the unigram `interact' is highly irrelevant for sentences, whereas
the bigram `interact with' is highly relevant. This may be because
all sentences in this corpus come from abstracts containing pharmacokinetic
DDI evidence (see ``Materials and Methods'' section). Thus, general
administration protocols and drug interaction terms are likely to
occur in the abstract as a whole but not necessarily in the evidence
sentences that actually report outcomes of the pharmacokinetic drug
interaction experiments. Similar patterns are observed in the more
extensive analysis provided in Supporting Information (section 2.2
in S1 Text), where relevant and irrelevant features are analyzed across
a wide range of classifier and feature transform configurations. There
we also show the results of a Principal Component Analysis of feature
weight coefficients chosen by different classifiers.

\begin{figure}[H]
\includegraphics[width=1\textwidth]{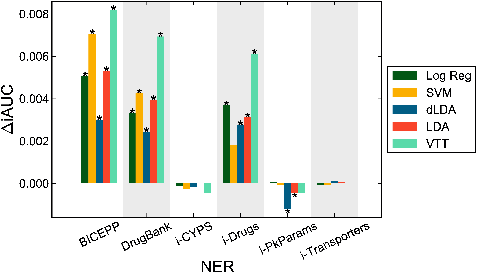}\protect\caption{\label{fig:NER-performance-avg-sent}\textbf{Performance impact of
sentence NER features}: Relative changes in iAUC scores on sentence
bigram runs (without transforms or dimensionality reductions) in combination
with different NER features. Significant changes (\emph{p}<0.05, two-tailed
test) in performance over respective classifiers without NER features
are indicated with asterisk `{*}'.}
\end{figure}

Finally, we tested the impact of additional features on sentence classification.
Though there is no metadata available in the sentence corpus, features
from NER tools can still be computed. Six NER features were tested:
\emph{BICEPP, DrugBank, i-CYPS, i-Drugs, i-PkParams, i-Transporters}
(see section ``Impact of NER and PubMed metadata on abstract classification''
for details). As before, we counted mentions of named biochemical
species and concepts specified by different NER tools in each sentence
and then included such counts as sentence features in addition to
the bigram and unigram textual features. Fig. \ref{fig:NER-performance-avg-sent}
shows relative iAUC changes when features from each of these NER tools
were included. Significant improvements (\emph{p}<0.05, two-tailed
test) above the corresponding classifier's performance without NER
features are indicated by an asterisk; performance according to MCC
and F1 measures is shown in Supporting Information (section 2.3 of
S1 Text). Notice that Naive Bayes was omitted since NER count features
are non-binary.

As in the abstract task, a few NER/Dictionary features improved performance
for several classifiers. The iAUC scores of nearly all classifiers
were significantly improved by three NER features: BICEPP, \emph{DrugBank},
and our internally developed \emph{i-Drugs} dictionary. These three
features represent counts of drugs names, showing that drug name counts
are helpful for classifying sentences as DDI-relevant vs. DDI-irrelevant.
Use of features from the BICEPP tool yielded the largest improvement
for every classifier. Table \ref{tab:senttable_NER} lists the performance
according to all measures for classifiers using the BICEPP features;
also listed are the corresponding best classifiers using only textual
features (bigram runs). The overall top classifiers (LDA and SVM)
showed significantly improved performance with the inclusion of these
NER features, reaching F1\ensuremath{\approx}0.76, MCC\ensuremath{\approx}0.65,
iAUC\ensuremath{\approx}0.83. In addition, VTT performance improved
significantly for all three measures with the inclusion of NER features.
Here VTT with bigrams performs better than other naive classifiers,
as expected given that this classifier was designed specifically to
handle such NER features \cite{lourencco2011linear,kolchinsky2010classification,abi2008uncovering}.
In contrast, dLDA (another naive classifier) did not benefit much
from the inclusion of NER features. 

\begin{table}[H]
\begin{centering}
\textsf{\footnotesize{}\begin{tabular}{l l l l l r}
\textbf{Classifier} & \textbf{Type} & \textbf{F1} & \textbf{MCC} & \textbf{iAUC} & \textbf{RP3} \\
\hline \hline
LDA & BICEPP & .757 (1) & .650 (1) & .831 (1) & 1 \\
\hline
SVM & BICEPP & .741 (4) & .639 (3) & .831 (1) & 12 \\
\hline
LDA & $-$ & .752 (2) & .642 (2) & .826 (4) & 16 \\
\hline
Log Reg & BICEPP & .738 (5) & .634 (4) & .828 (3) & 60 \\
\hline
VTT & BICEPP & .742 (3) & .629 (7) & .805 (7) & 147 \\
\hline
SVM & $-$ & .736 (6) & .633 (5) & .824 (5) & 150 \\
\hline
Log Reg & $-$ & .734 (8) & .630 (6) & .823 (6) & 288 \\
\hline
VTT & $-$ & .736 (6) & .617 (8) & .797 (8) & 432 \\
\hline
Naive Bayes & $-$ & .734 (8) & .619 (8) & .796 (10) & 640 \\
\hline
dLDA & BICEPP & .711 (10) & .603 (10) & .797 (8) & 800 \\
\hline
dLDA & $-$ & .710 (11) & .600 (11) & .794 (11) & 1331 \\
\hline
\end{tabular}}
\par\end{centering}{\footnotesize \par}

\medskip{}
\protect\caption{\label{tab:senttable_NER}\textbf{Sentence classification performance
with NER features}: Performance of different sentence classifiers
with the count features obtained via the BICEPP NER tool; also listed
are the corresponding best classifiers using only textual features
(bigram runs; indicated by -). F1, MCC, and iAUC performance measures
are listed; the rank of the classifiers according to each measure
is reported in parenthesis in the respective column. Classifiers are
ordered according to the rank product (RP3) of the three measures
(last column).}
\end{table}

\section*{Discussion}

We have demonstrated that current BLM methods for text classification
can reliably identify PubMed abstracts containing pharmacokinetic
evidence of drug-drug interactions, as well as extract specific sentences
that mention such evidence from relevant abstracts. The performance
reached on a corpus of carefully annotated pharmacokinetics literature
is quite high for both abstract classification (reaching F1\ensuremath{\approx}0.93,
MCC\ensuremath{\approx}0.74, iAUC\ensuremath{\approx}0.99) and evidence
sentence extraction (F1\ensuremath{\approx}0.76, MCC\ensuremath{\approx}0.65,
iAUC\ensuremath{\approx}0.83). To explore the capability of BLM in
the pharmacokinetics DDI context, where there are no existing directly-relevant
corpora or experiments, we pursued a thorough comparison of the performance
of several linear classifiers using different combinations of unigrams,
bigrams, PubMed metadata, and NER features. We also tested the effects
of applying feature transforms and dimensionality reduction.

From a classification performance perspective, some results are noteworthy:
in terms of textual features, bigrams in combination with unigrams
performed significantly better than unigrams alone. However, performance
in unigram versus bigram runs for the same classifier differed by
no more than one percent for iAUC and MCC. Thus, while bigram features
did contain some additional information about class membership, the
amount of this information was not large. 

In our experiments, feature transforms and PCA-based dimensionality
reduction significantly improved performance for several classifiers
(especially ``naive'' classifiers such as dLDA, which assume feature
independence), but did not significantly improve the overall best
performance. We also found that a sophisticated version of the LDA
classifier dominated performance in both the abstract and sentence
tasks. This classifier used SVD to eliminate rank-deficiency in the
feature occurrence matrices and performed shrinkage of the feature
covariance matrix for regularization (see ``Materials and Methods'').

From the drug-interaction domain perspective, feature analysis in
the abstract task revealed that pharmacokinetic DDI evidence in the
literature is highly correlated with terms that explicitly indicate
interaction (including MeSH terms), enzyme inhibitors (including substance
names via the RN metadata field in PubMed), DDI administration protocols,
and study design. At the sentence level, drug interaction evidence
from a pharmacokinetics perspective is highly correlated with terms
that express experimental results, such as numerical values, measures
of drug clearance, expressions of quantitative changes, as well as
adverbs expressing significance of evidence. Feature analysis at the
abstract level also revealed that lack of DDI evidence in the pharmacokinetics
literature (irrelevant class) is highly correlated with some terms
from PubMed metadata fields, as well as those pertaining to genomic
or general medical terminology. At the sentence level, sentences in
relevant abstracts but without DDI evidence tend to include terminology
relevant to pharmacokinetics protocols, as well as more generic interaction
discourse or biomedical concepts.

Since many important features came from PubMed metadata fields, we
looked at changes in iAUC scores when features from different PubMed
metadata fields were omitted. We found that only the omission of MeSH
terms significantly affected abstract classification performance.
Nonetheless, while statistically significant, the drop in performance
was rather small (affecting only millesimals of the iAUC, iAUC\ensuremath{\approx}0.98
without), indicating that abstract classification does not depend
strongly on the inclusion of MeSH term features. This is an important
consideration since MeSH terms may not be immediately added to publications,
with statistics indicating that only 50\% of citations are annotated
within 60 days of inclusion in PubMed \cite{huang2011recommending}.
Therefore, classification and evidence extraction from brand new articles
should not rely on such metadata.

We also tested the effect of including features extracted using named
entity recognition (NER) and dictionary tools, namely those for detecting
possibly-relevant chemical, genomic, metabolomic, drug, and pharmacokinetic
entities. Generally, dictionaries like BICEPP, \emph{i-Drugs}, and
DrugBank, which counted the number of times drug names appeared, significantly
improved performance for several classifiers on both the abstract
classification and evidence sentence extraction tasks (an exception
to this was the lack of improvement on abstracts when including DrugBank
features, an effect that needs further investigation). Nonetheless,
as for MESH term features in abstract classification, the resulting
performance increases were modest, even if statistically significant.
This again demonstrates that relevant-class information can be extracted
from abstracts and sentences using solely the statistics of unigram
and bigram textual features.

Notably, relevant and irrelevant documents and sentences both derive
from the pharmacokinetics literature and therefore share similar feature
statistics. This makes distinguishing between them a nontrivial text
classification problem, though also a more practically relevant one
(e.g. for a researcher who needs to automatically label potentially
relevant documents retrieved from PubMed). Nonetheless, several classifiers
reached high performance; for example, the abstract ranking performance
(iAUC\ensuremath{\approx}0.99) has little room for further improvement,
though the classification performance --- while high for this type
of problem --- can still be improved. 

We observed that many different pipeline configurations reached near-optimal
performance. Even though some performance differences between configurations
were statistically significant, they were small. For instance, iAUC
differences between best and worst classifiers varied by no more than
1 percent in the abstract task and 5 percent in the sentence task.
This demonstrates that classification performance in our experiments
was robust to the classifier utilized, and that a BLM pipeline for
this problem would do similarly well independently of classifier chosen.
In particular, while ``non-naive'' classifiers (which consider feature
covariances) performed better than naive classifiers, the latter are
still capable of competitive performance. These results suggest a
fundamental limit on the amount of statistical signal present in the
labels and feature distributions of the corpora as extractable by
linear classifiers. However, it is worth noticing that an analysis
of both abstract- and sentence-trained feature weight coefficients
shows systematic differences between weights selected by naive and
non-naive classifiers (see Supporting Information, S1 Text), indicating
that different classifiers emphasize distinct semantic features. Furthermore,
it is possible that performance could be improved by the use of non-linear
classifiers or features produced by more finely DDI-tuned NER tools,
relation extraction or NLP methods, or other sophisticated feature-generation
techniques. Indeed, the larger performance variation observed in the
sentence task suggests that sentence extraction performance may improve
with larger amounts of training data (which would permit better estimates
of feature covariances).

It is not trivial to compare our performance results with those previously
reported in the literature. First, there is no gold standard for DDI
evidence sentence extraction, especially for a specific evidence-type
such as pharmacokinetics. Second, most sentence extraction tasks in
the biomedical domain involve extraction of passages which can contain
several sentences (e.g. the protein-protein interaction subtask in
Biocreative II) or passages relevant for a set of specific targets
(e.g. Gene Ontology annotations for specific gene names in Biocreative
I\cite{hirschman2005overview} and IV\cite{mao2013gene}). Due to
these difficulties, the performance on those tasks has been comparatively
low, e.g. in BioCreative IV the best F1 score in the gene ontology
evidence extraction task was 0.27\cite{mao2013gene} (in Biocreative
II, due to possible overlap and multiple accepted passages, the preferred
performance measure was the mean reciprocal rank which reached 0.87
\cite{krallinger2008overview,abi2008uncovering}). Considering that
our performance on the sentence task is higher than what is typically
reported for the abstract classification in the biomedical domain
(e.g. PPI abstract classification in the BioCreative Challenge III
reached F1\ensuremath{\approx}0.61, MCC\ensuremath{\approx}0.55, iAUC\ensuremath{\approx}0.68
\cite{krallinger2011protein}), the classifiers trained on our sentence
corpus reached a very good level of performance, indicating that the
corpus is well annotated and that the task is highly feasible. Given
the performance of our approach in extracting pharmacokinetic evidence,
the classification methodology and associated corpus may be useful
in the previously explored task of extracting interacting drug pairs
from the literature. For example, it may be more effective to first
identify DDI sentences containing specific types of evidence and then
extract the interacting drug names from them, using automated methods
or human expertise tailored to that specific type of DDI evidence. 

To conclude, we provide a thorough report of the capability of linear
classifiers to automatically extract pharmacokinetics evidence of
DDI from an abstract- and sentence-level annotated corpus. Given the
high performance observed on both abstract and sentence classification
for all classifiers, including the simplest ones, we conclude that
under realistic classification scenarios automatic BLM techniques
can identify PubMed abstracts reporting DDI backed by pharmacokinetic
evidence, as well as extract evidence sentences from relevant abstracts.
These results are important because pharmacokinetic evidence can be
essential in identifying causal mechanics of putative DDI and as input
for further pharmacological and pharmaco-epidemiology investigation.
More generally, our work shows that BLM can be safely included in
DDI discovery pipelines where attention to distinct types of evidence
is necessary. In future work, we intend to use our methodology to
mine large corpora for both pharmacokinetic and other types of DDI
experimental evidence. Such evidence can help fill knowledge gaps
that exist in the DDI domain, with the ultimate goal of reducing the
incidence of adverse drug reactions and contributing to the development
of alternative safe treatments.

\section*{Acknowledgments}

The authors thank Shreyas Karnik for help with the preparation of
the corpora.

\bibliographystyle{plos2015}
\bibliography{writeup}



\section*{Supplementary material}

# Extraction of Pharmacokinetic Evidence of Drug-drug Interactions from the Literature: SUPPORTING INFORMATION

*Artemy Kolchinsky, Anália Lourenço, Heng-Yi Wu, Lang Li, Luis M. Rocha*

## 1 Abstract performance

The following table lists classification performance according to F1, MCC, iAUC, and Accuracy measures for different classifiers on unigram and bigram abstract runs. It includes the VTTcv classifier (VTT with a cross-validated threshold), which is not discussed in the main article.

| Classifier | Type | F1 | MCC | iAUC | Accuracy |
|---|---|---|---|---|---|
| Log Reg | Bigram | .929 | .698 | .984 | .891 |
| Log Reg | Unigram | .927 | .689 | .980 | .888 |
| Naive Bayes | Bigram | .920 | .661 | .981 | .878 |
| Naive Bayes | Unigram | .919 | .672 | .978 | .878 |
| SVM | Bigram | .928 | .693 | .983 | .890 |
| SVM | Unigram | .927 | .689 | .980 | .888 |
| VTTcv | Bigram | .916 | .697 | .980 | .877 |
| VTTcv | Unigram | .911 | .663 | .977 | .868 |
| dLDA | Bigram | .909 | .670 | .975 | .868 |
| dLDA | Unigram | .908 | .658 | .974 | .865 |
| LDA | Bigram | .931 | .728 | .984 | .897 |
| LDA | Unigram | .926 | .719 | .983 | .891 |
| VTTcv | Bigram | .916 | .697 | .980 | .877 |
| VTTcv | Unigram | .911 | .663 | .977 | .868 |
| VTT | Bigram | .919 | .656 | .980 | .876 |
| VTT | Unigram | .918 | .662 | .977 | .876 |

## 1.1 Abstract performance: Feature transforms and dimensionality reduction

The following charts show iAUC, F1, and MCC performance on abstract bigram runs when feature transforms and dimensionality reductions are applied. Stars indicate configurations that performed significantly better ($p<0.05$, one-tailed test) than the no-transform, no-dimensionality-reduction configuration of the same classifier (indicated with larger circles). Naive Bayes was not tested since it is only applicable to binary data, and VTT was only tested on the sparse transforms (i.e., without PCA-based dimensionality reduction).

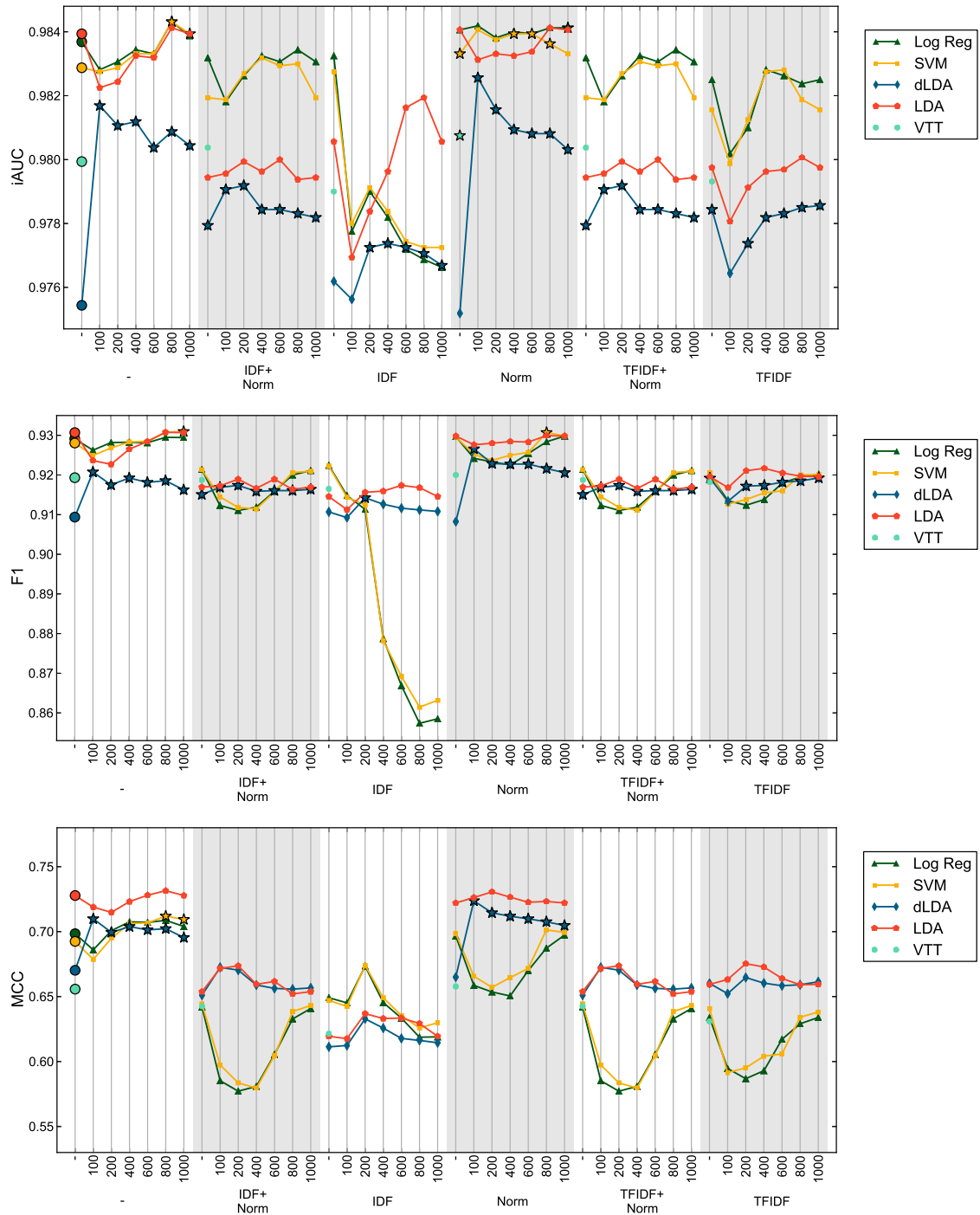

## 1.2 Abstract Performance: Most relevant and irrelevant features

A linear classifier separates classes using a hyperplane defined by a set of feature coefficients. The impact of a given feature on classification is naturally quantified by the sign and amplitude of its hyperplane coefficient. The increase of the value of a feature with a large positive (negative) coefficient produces a large increase in a document's propensity to be classified as relevant (irrelevant). Coefficients for different features are made comparable by an appropriate normalization: we multiply each feature's coefficient by the standard deviation of the feature's values in the training data, producing what is referred to as a 'standardized coefficient' in the linear regression literature. For the abstract bigram runs, the following figure shows the ranks of the most relevant and the reverse rank of the most irrelevant features. *RELEVANT FEATURES* includes any feature whose standardized coefficient was among the top 5 most positive standardized coefficients for any transform/classifier combination, while *IRRELEVANT FEATURES* includes any feature whose standardized coefficient was among the top 5 most negative standardized coefficients for any transform/classifier combination. Transforms are organized in the vertical columns, while classifiers are distinguished by color and marker style. For relevant (irrelevant) features, markers are positioned according to their rank (reverse rank) among the most positive (negative) features for a given classifier and transform combination.

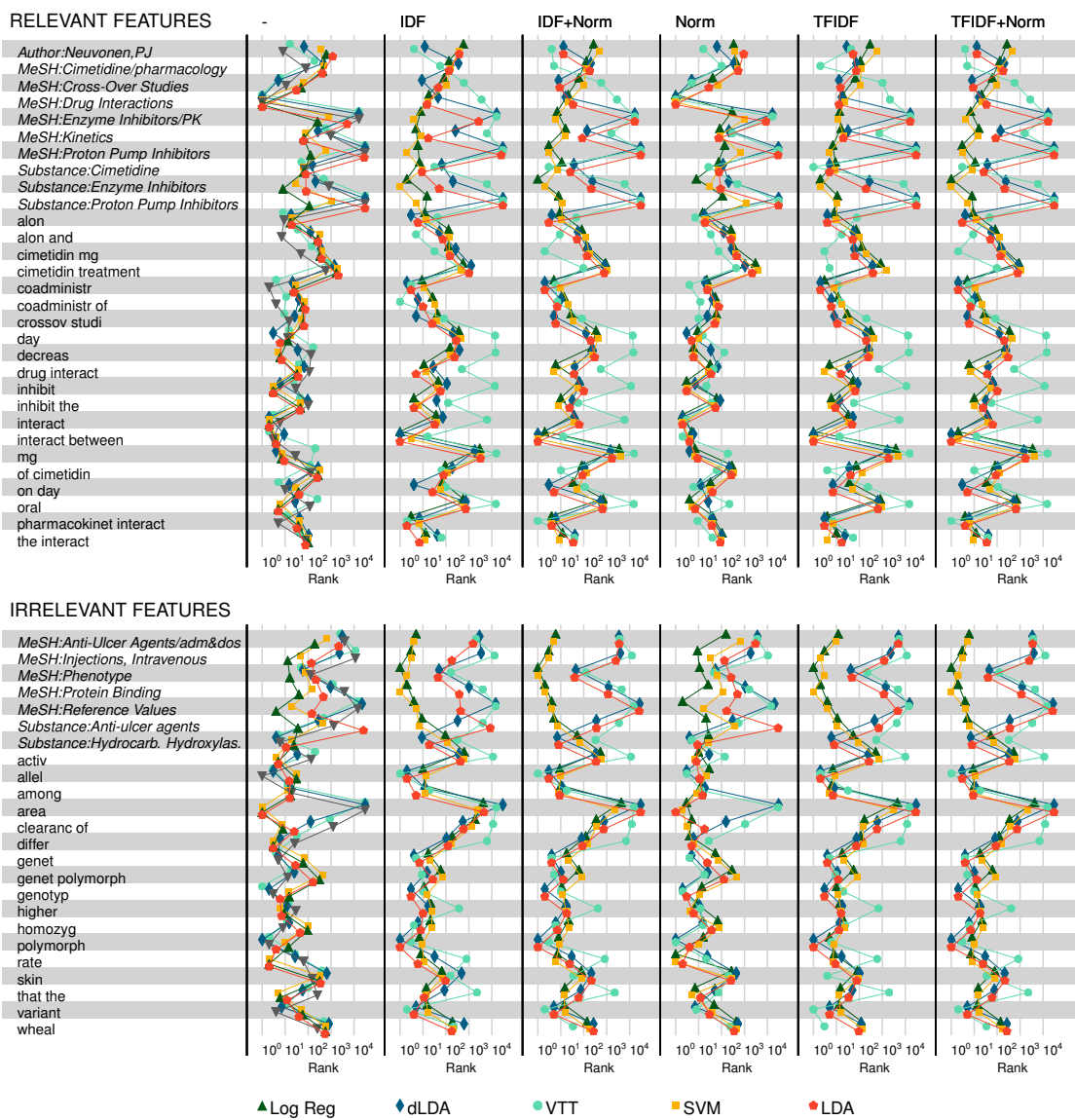

To further compare the importance that different classifier/transform combinations gave to different features, we performed principal component analysis on a matrix composed of the hyperplane coefficients of all such combinations. The following figure shows each transform-classifier hyperplane in terms of its loading on the first two principal components (PCs). This projection separates classifiers that use feature covariance information (LDA, SVM, and Logistic Regression) and those that don't (Naive Bayes, dLDA, VTT). It also groups configurations according to feature transforms, with configurations that included IDF-like transforms clustering separately from those that used no transforms or a simple L2-normalization. In general, SVM and Logistic Regression produce very similar feature loadings, likely due to the fact that they optimize similar cost functions during training.

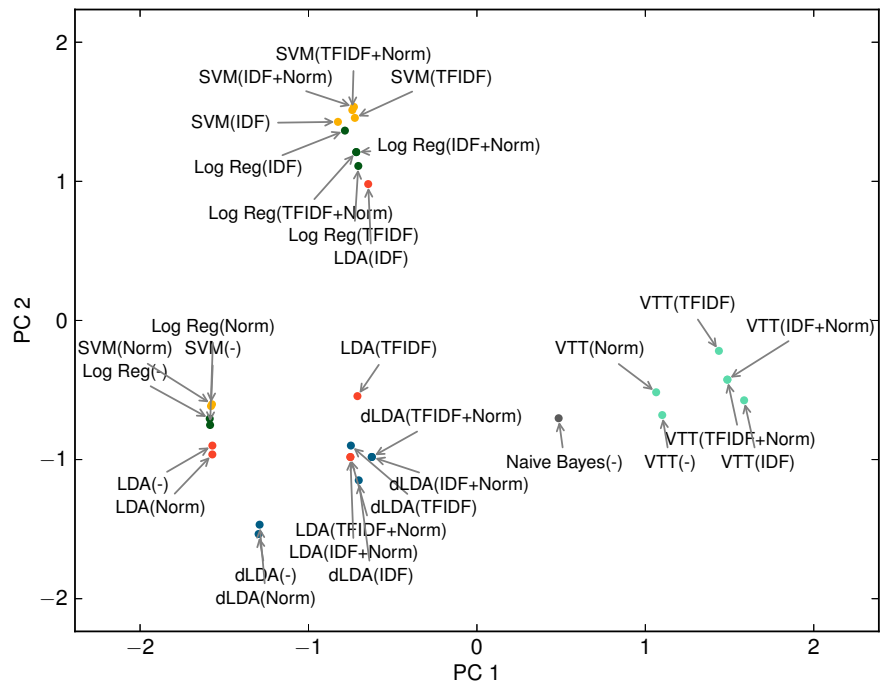

## 1.3 Abstract Performance: NER and Metadata Features

The following figures plot the relative changes in F1 and MCC performance when including vs. not-including metadata and NER-derived features on abstract bigram runs with feature transforms. Significant changes ($p<0.05$, two-tailed test) are indicated by asterisks. For metadata, changes in performance are measured while still including features from the other 4 metadata fields.

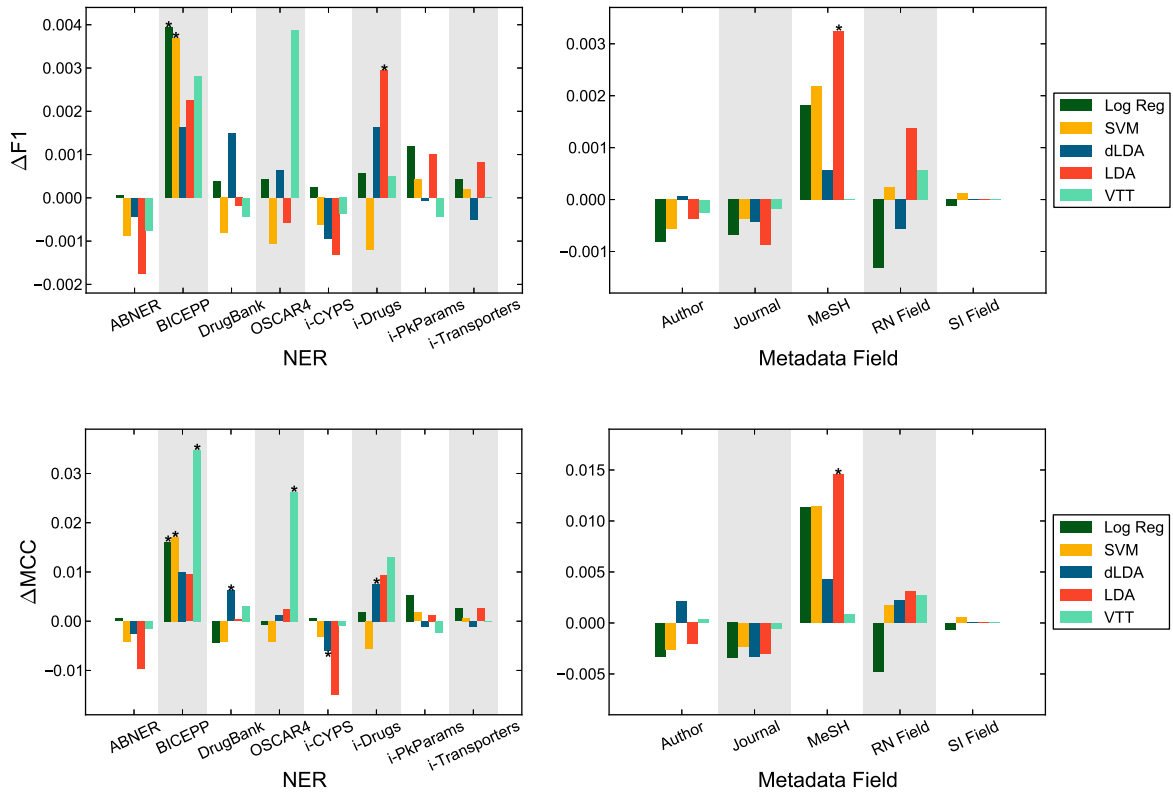

## 2 Sentence performance

The following table lists classification performance according to F1, MCC, iAUC, and Accuracy measures for different classifiers on unigram and bigram sentence runs. It includes the VTTcv classifier (VTT with a cross-validated threshold), which is not discussed in the main article.

| Classifier | Type | F1 | MCC | iAUC | Accuracy |
|---|---|---|---|---|---|
| Log Reg | Bigram | .734 | .630 | .823 | .848 |
| Log Reg | Unigram | .734 | .629 | .818 | .847 |
| Naive Bayes | Unigram | .736 | .617 | .791 | .835 |
| Naive Bayes | Bigram | .734 | .619 | .796 | .839 |
| SVM | Unigram | .735 | .630 | .819 | .847 |
| SVM | Bigram | .736 | .633 | .824 | .848 |
| VTTcv | Bigram | .733 | .608 | .797 | .822 |
| VTTcv | Unigram | .729 | .608 | .789 | .831 |
| dLDA | Bigram | .710 | .600 | .794 | .836 |
| dLDA | Unigram | .732 | .613 | .790 | .834 |
| LDA | Bigram | .752 | .642 | .826 | .848 |
| LDA | Unigram | .750 | .636 | .819 | .843 |
| VTTcv | Bigram | .733 | .608 | .797 | .822 |
| VTTcv | Unigram | .729 | .608 | .789 | .831 |
| VTT | Bigram | .736 | .617 | .797 | .836 |
| VTT | Unigram | .733 | .606 | .789 | .822 |

## 2.1  Sentence performance: Feature transforms and dimensionality reduction

The following charts show the F1 and MCC performance performance on bigram runs when feature transforms and dimensionality reductions are applied. Significant improvements ($p<0.05$, one-tailed test) compared to the same classifier applied to non-transformed data are indicated by stars.

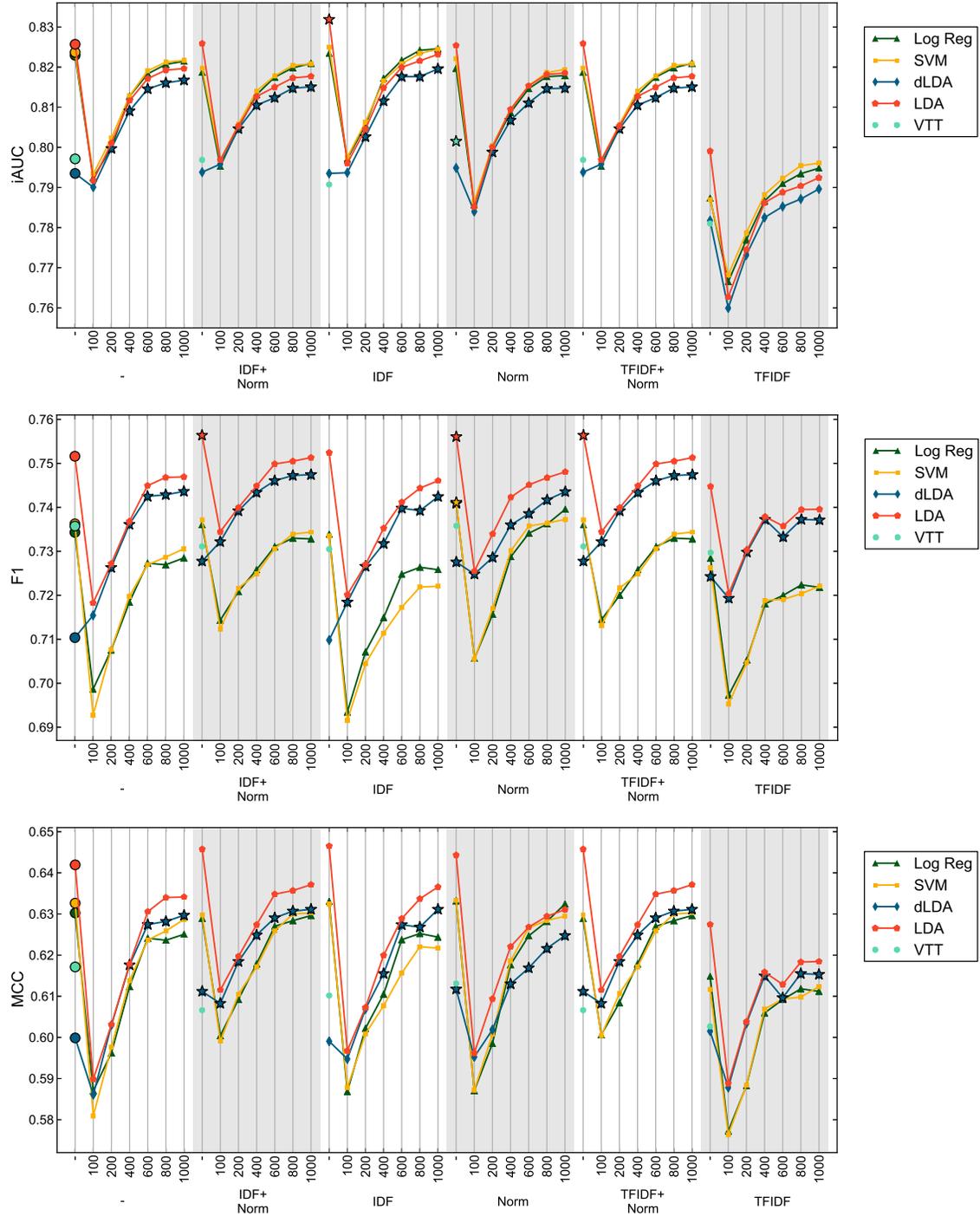

## 2.2    Sentence Performance: Most relevant and irrelevant features

We analyzed which features were most relevant and irrelevant for identifying evidence sentences using the same methodology as for abstracts (described in section 1.2). In the following figure, *RELEVANT FEA-TURES* includes any feature whose standardized coefficient was among the top 5 most positive standardized coefficients for any transform/classifier combination, while *IRRELEVANT FEATURES* includes any whose standardized coefficient was among the top 5 most negative standardized coefficients for any transform/classifier combination. Transforms are organized in the vertical columns, while classifiers are distinguished by color and marker style. For relevant (irrelevant) features, markers are positioned according to their rank (reverse rank) among the most positive (negative) features for a given classifier and transform combination.

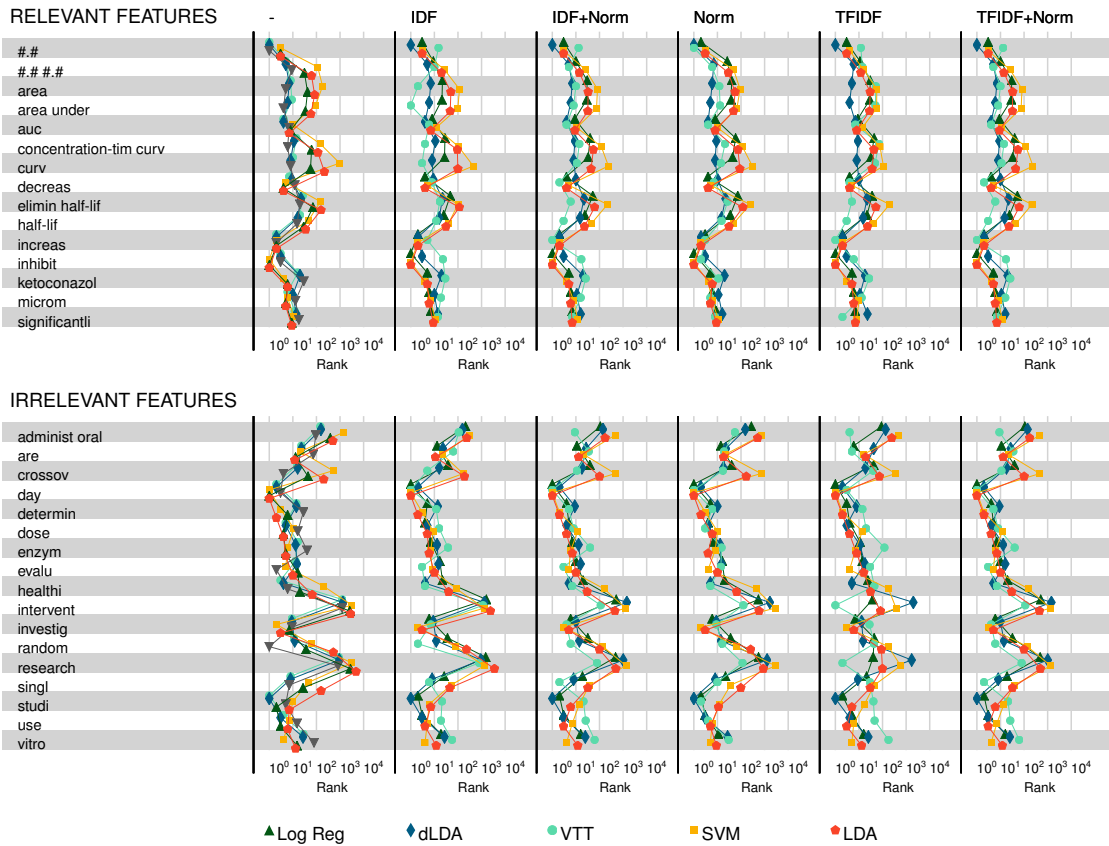

As in section 1.2, we perform principal component analysis of separating hyperplanes produced by different transforms and classifiers trained on the sentence corpus. Hyperplanes are generally grouped by classifier in this projection, with those corresponding to 'naive' classifiers that do not use feature covariances (VTT, dLDA, Naive Bayes) clustering separately from those that do (LDA, SVM, Logistic Regression).

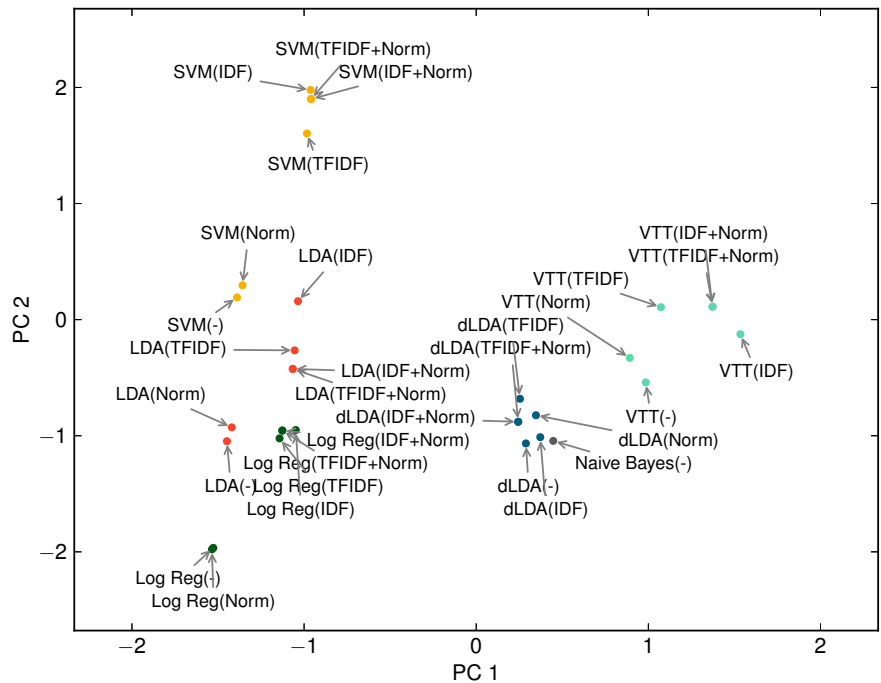

## 2.3   Sentence Performance: Impact of NER Features

The following figures plot the relative changes in F1 and MCC performance when including vs. not-including NER-derived features on non-transformed sentence bigram runs. Significant changes (p<0.05, two-tailed test) indicated by asterisks.

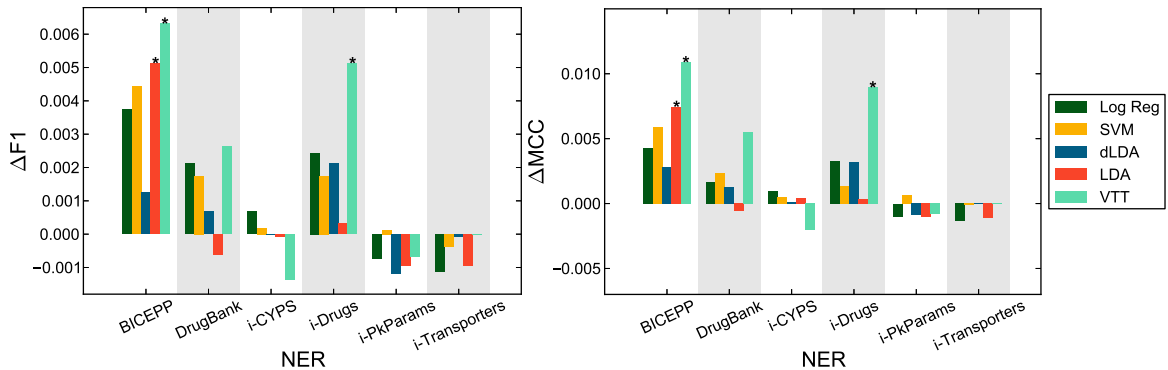



\end{document}